\documentclass[10pt]{article} 
\usepackage[preprint]{tmlr}

\usepackage{amsmath,amsfonts,bm}

\def\eqref#1{equation~\ref{#1}}

\def\1{\bm{1}}

\DeclareMathAlphabet{\mathsfit}{\encodingdefault}{\sfdefault}{m}{sl}
\SetMathAlphabet{\mathsfit}{bold}{\encodingdefault}{\sfdefault}{bx}{n}

\usepackage{hyperref}
\usepackage{url}
\usepackage{booktabs}
\usepackage{fontawesome5}
\usepackage{graphicx}
\usepackage{caption}
\usepackage{array}
\usepackage{xcolor}
\usepackage{titlesec}
\definecolor{linkblue}{HTML}{2A64B9}
\titleformat*{\section}{\color{blue!50!black}\Large\bfseries}
\titleformat*{\subsection}{\color{blue!50!black}\large\bfseries}
\titleformat*{\subsubsection}{\color{blue!50!black}\normalsize\bfseries}
\usepackage{tikz}
\usepackage{forest}

\useforestlibrary{edges}

\title{Reinforcement Learning in GUI Agents:\\
A Survey Toward Digital Inhabitants}
\makeatletter
\renewcommand{\headrule}{\vspace*{3pt}\hrule\@height\headrulewidth\@width\headwidth\vskip-\headrulewidth}
\makeatother

\author{%
    Junan Hu$^{1,2}$,
    Jian Liu$^{2\ddagger}$,
    Jingxiang Lai$^{2}$,
    Jiarui Hu$^{3}$,
    Yiwei Sheng$^{4}$,
    Shuang Chen$^{5}$,
    \\
    Jian Li$^{5}$,
    Dazhao Du$^{2}$,
    Song Guo$^{2*}$
    \vspace{1mm} \\
    {\small $^1$Shandong University \quad
    $^2$The Hong Kong University of Science and Technology \quad
    $^3$The Hong Kong University} \\
    {\small $^4$Shanghai Jiao Tong University \quad
    $^5$Tencent} \\
    \vspace{0.5mm}
    {\small $\ddagger$ Project lead \qquad
    $^{*}$ Corresponding author} \\
    \vspace{0.5mm}
    {\small \faEnvelope[regular]\ \href{mailto:junanhu06@gmail.com}{\textcolor{linkblue}{junanhu06@gmail.com}} \,
    \href{mailto:jliuhr@connect.ust.hk}{\textcolor{linkblue}{jliuhr@connect.ust.hk}} \qquad
    \faGithub\ \href{https://github.com/Steve2457/Awesome-RL-GUI-Agents}{\textcolor{linkblue}{Awesome-RL-GUI-Agents}}}
}

\long\def\rev#1{{#1}}

\def\month{MM}  
\def\year{YYYY} 
\def\openreview{\url{https://openreview.net/forum?id=XXXX}} 

\makeatletter
\renewcommand{\@maketitle}{%
\vbox{\hsize\textwidth
\vspace*{-0.55in}
{\centering\LARGE\bfseries\rmfamily\color{blue!50!black} \@title\par}%
\vskip 0.06in
{\centering \@author\par}%
\vskip 0pt}}
\makeatother

\begin{document}

\maketitle
\thispagestyle{empty}

\makeatletter
\renewenvironment{abstract}{\vskip 0.01in\noindent\ignorespaces}{\vskip 0.01in}
\makeatother


\begin{abstract}
\textbf{Abstract:} Graphical User Interface (GUI) agents have emerged as a promising paradigm for intelligent systems that perceive and interact with graphical interfaces visually. Yet supervised fine-tuning alone cannot handle long-horizon credit assignment, distribution shifts, and safe exploration in irreversible environments, making Reinforcement Learning (RL) a central methodology for advancing automation. In this work, we present \rev{a comprehensive RL-centered} overview of the intersection between RL and GUI agents, and examine how this research direction may evolve toward digital inhabitants. We propose a principled taxonomy that organizes existing methods into Offline RL, Online RL, and Hybrid Strategies, and complement it with analyses of reward engineering, data efficiency, and key technical innovations. Our analysis reveals several emerging trends: the tension between reliability and scalability is motivating the adoption of composite, multi-tier reward architectures; GUI I/O latency bottlenecks are accelerating the shift toward world-model-based training, which can yield substantial performance gains; and \rev{early evidence of deliberation-like traces suggests that structured rewards may reduce reliance on explicit reasoning annotations in some settings.} We distill these findings into a roadmap covering process rewards, continual RL, cognitive architectures, and safe deployment, aiming to guide the next generation of robust GUI automation and its agent-native infrastructure.
\end{abstract}

\vspace{-0.2em}

\begin{center}
    \includegraphics[width=0.89\textwidth]{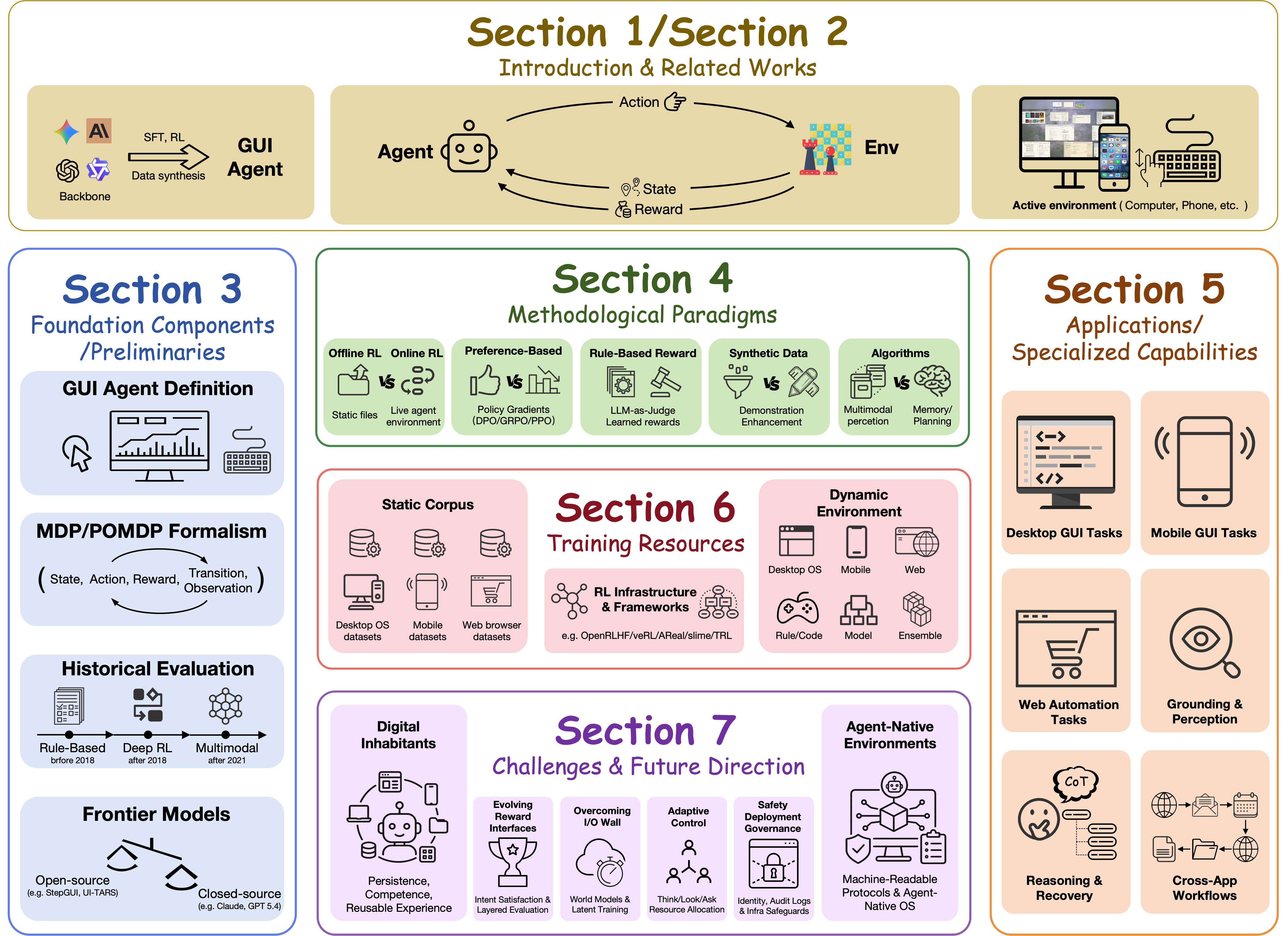}
    \captionof{figure}{Overview of the survey structure. Sections~\ref{sec:introduction} and~\ref{sec:related-works} introduce the background of GUI agent and related works; Section~\ref{sec:preliminaries} presents the preliminaries; Section~\ref{sec:rl-methods} reviews RL methodologies; Section~\ref{sec:key-dimensions} summarizes key dimensions and applications; Section~\ref{sec:training-resources} surveys training resources; and Section~\ref{sec:future} discusses challenges and future directions.}
    \label{fig:section}
\end{center}

\newpage
\tableofcontents
\newpage

\section{Introduction}
\label{sec:introduction}

GUI agents---intelligent systems that perceive graphical user interfaces visually and execute tasks through simulated human inputs such as mouse clicks, typing, and scrolling---sit at the frontier of a paradigm shift in AI: from passive information processing to active task execution in real digital environments. Unlike traditional automation tools such as Selenium~\citep{Selenium_2023} or Robotic Process Automation (RPA)~\citep{van2018robotic}, which depend on brittle DOM selectors or rigid coordinate scripts, modern GUI agents leverage visual perception and semantic reasoning to generalize across heterogeneous operating systems, dynamic web pages, and previously unseen applications. Sequential decisions, delayed feedback, and distribution shift in these environments motivate \textbf{Reinforcement Learning (RL)} as one approach to training GUI agents when suitable rewards and safe interaction are available.

\paragraph{Why RL? A formal argument.}
Consider a GUI task formalized as a finite-horizon Partially Observable Markov Decision Process \rev{$\mathcal{M} = (\mathcal{S}, \mathcal{A}, \mathcal{O}, \mathcal{T}, \mathcal{Z}, \mathcal{R}, \gamma, H)$}. The latent state $s_t \in \mathcal{S}$ represents the underlying digital environment, while the agent receives an observation $o_t \in \mathcal{O}$, such as a screenshot, OCR text, DOM snapshot, or accessibility tree, generated by the observation model $\mathcal{Z}(o_t \mid s_t, a_{t-1})$. The agent executes an action $a_t \in \mathcal{A}$ (e.g., $\texttt{click}(x,y)$, $\texttt{type}(\text{``query''})$), and receives a reward $r_t = \mathcal{R}(s_t, a_t)$ or an observation-based proxy. Policies are therefore history- or belief-conditioned, e.g., $\pi(a_t \mid \mathcal{H}_t)$ or $\pi(a_t \mid b_t)$, where $\mathcal{H}_t=(o_0,a_0,\ldots,o_t)$. In realistic GUI environments, the following three properties motivate RL-based optimization as a complement to supervised fine-tuning (SFT) when suitable rewards and safe interaction are available:

\begin{enumerate}
    \item \textbf{Sparse, delayed rewards.} The reward signal is typically binary---$r_T = +1$ upon task completion, $r_t = 0$ for all intermediate steps $t < T$---with trajectory lengths $T$ ranging from 50 to 200 steps. No intermediate supervision is available for sub-step correctness. This creates a severe credit assignment problem: $\nabla_\theta J(\pi) = \mathbb{E}_\pi\left[\sum_{t=0}^{\rev{H}} \nabla_\theta \log \pi_\theta(a_t|\rev{\mathcal{H}_t}) \cdot \hat{A}_t\right]$, where the advantage $\hat{A}_t$ must propagate reward information across dozens of \rev{partially observed} steps.

    \item \textbf{Distribution shift and environment non-stationarity.} Static demonstration datasets capture specific UI versions, but real interfaces evolve continuously through updates, A/B testing, and redesigns. Behavioral cloning from such data suffers from compounding covariate shift: \rev{under a per-step imitation error $\epsilon$, the expected number of mistakes over horizon $H$ scales as $\mathcal{O}(H\epsilon)$, while worst-case cumulative cost degradation under distribution shift can scale as $\mathcal{O}(H^2\epsilon)$~\citep{ross2011reduction}.}

    \item \textbf{Optimization beyond human imitation.} RL enables agents to discover execution paths more efficient than \rev{available} demonstrations. \rev{Behavioral cloning is constrained by the quality and coverage of the demonstration distribution and does not directly optimize deployment-time reward. When demonstrations are noisy, redundant, or suboptimal, RL with reliable rewards and safe exploration can improve over demonstrated behavior.}
\end{enumerate}

\paragraph{Why now? GUI as the ultimate RL laboratory.}
Beyond addressing the aforementioned structural challenges, the current acceleration of RL in GUI agents is driven by a fundamental property distinguishing them from pure text LLMs: \textbf{verifiability}. In plain text generation, defining an ``absolute correct'' response is notoriously elusive, rendering reward modeling subjective and prone to hallucination or reward hacking. \rev{GUI environments, by contrast, often provide more objective task evidence}---a successfully navigated URL, a predictably altered DOM tree, or a confirmed database transaction. \rev{This structured feedback makes GUI environments a natural testbed for Reinforcement Learning with Verifiable Rewards (RLVR), while still leaving open challenges for ambiguous goals, partial observability, and reward hacking.} Consequently, applying RL here transcends being a mere ``patch solution'' for imitation learning shortcomings; it establishes a \rev{closed-loop path toward agents that} can autonomously interact, receive \rev{auditable} environmental feedback, and continuously \rev{self-improve under constraints}. More broadly, GUI agents may serve as the transitional substrate through which AI systems evolve from task-specific operators of human software into persistent digital inhabitants, while revealing what future agent-native infrastructure should look like.

\paragraph{RL's proven track record.}
The case for RL in GUI automation builds on a series of validated breakthroughs. AlphaGo~\citep{silver2016mastering} and AlphaZero~\citep{silver2017mastering} demonstrated that RL with binary win/loss rewards could surpass human experts through self-play---precisely the regime of sparse terminal feedback that GUI tasks inhabit. RLHF~\citep{ouyang2022training} and DPO~\citep{rafailov2023direct} showed that RL can optimize for objectives too nuanced for explicit labels, enabling GPT-4~\citep{achiam2023gpt} and Claude~\citep{bai2022constitutional} to align with complex human preferences. Most recently, reasoning-focused models like OpenAI o1~\citep{jaech2024openai} and DeepSeek-R1~\citep{guo2025deepseek} employed RL with Verifiable Rewards (RLVR) to achieve breakthrough performance on mathematical and multi-step reasoning tasks. GUI agents share all the characteristics that made RL successful in these domains: sequential decision-making under uncertainty, sparse delayed feedback, and the need to generalize beyond static training distributions.

\paragraph{GUI agents vs. CLI agents: The necessity of visual interaction.}
While command-line interface (CLI) agents excel in efficiency and determinism for expert-oriented workflows---where structured text streams and explicit exit codes provide unambiguous feedback~\citep{yang2023intercode, gandhi2026endless}---they face fundamental coverage limitations. CLI agents operate exclusively within systems exposing programmatic APIs or terminal interfaces, typically constraining them to software engineering, system administration, and database operations~\citep{bechard2026terminal, li2026skillsbench}. In contrast, the vast majority of real-world digital systems remain inaccessible via APIs: legacy enterprise software in finance and healthcare, proprietary industrial control systems, and countless closed-source applications built atop graphical interfaces without programmatic access~\citep{van2018robotic}. 

\begin{figure}[h]
    \centering
    \includegraphics[width=0.93\textwidth]{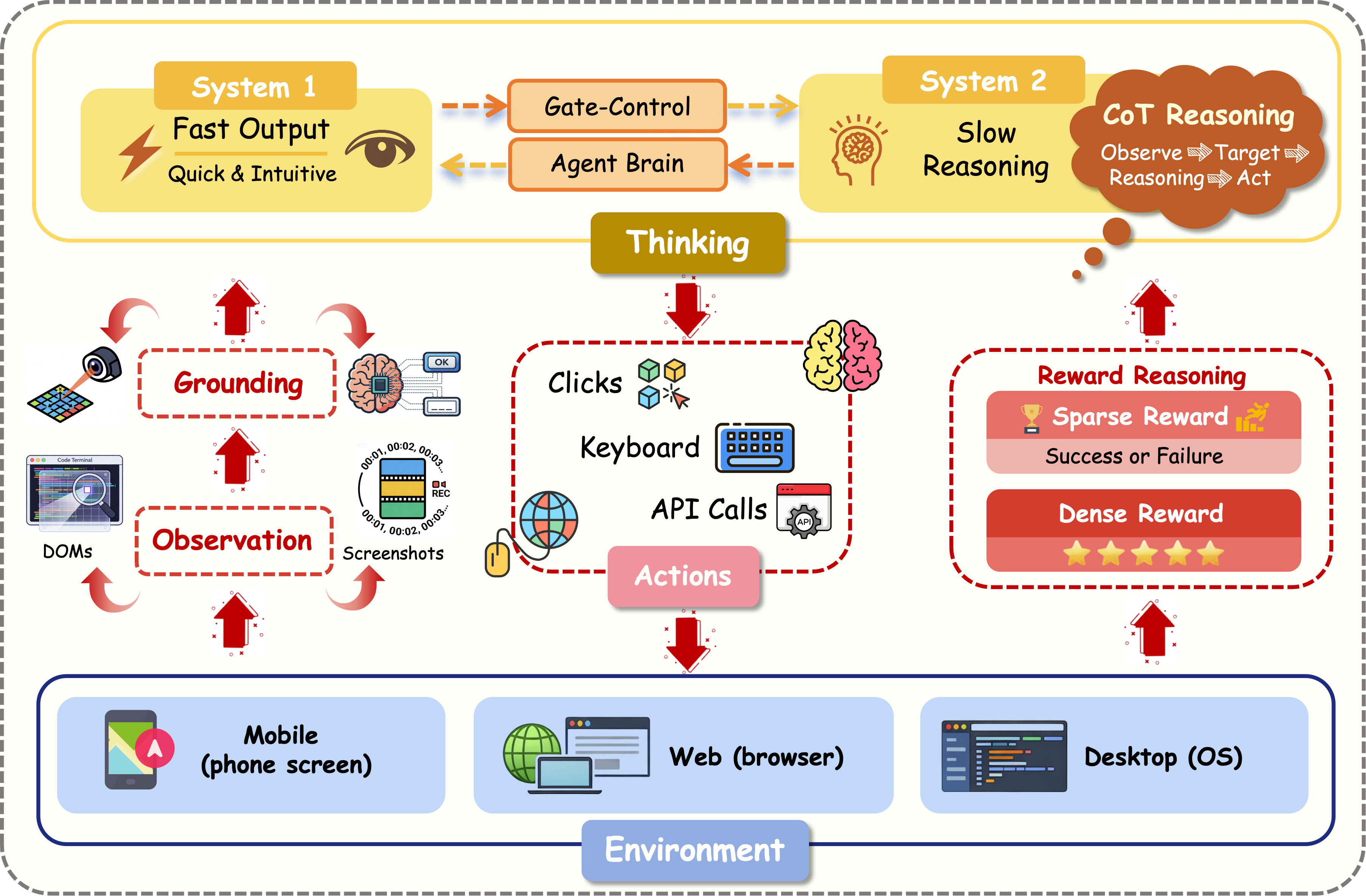}
    \caption{Overview of the RL training pipeline for GUI agents. The agent perceives the GUI environment through screenshots, reasons about the task, and executes actions. RL optimizes the policy through reward signals derived from task completion, visual grounding accuracy, and intermediate reasoning quality.}
    \label{fig:gui-agent-rl-overview}
\end{figure}

GUI agents address this \textbf{long-tail coverage problem} through visual perception. By reasoning over pixel-level observations or accessibility trees, they can interact with \emph{any} system a human can operate---regardless of whether backend APIs exist. Modern unified frameworks like ClawGUI~\citep{tang2026clawgui} demonstrate this universality across diverse platforms and applications. However, this generality comes at a cost: GUI environments exhibit higher observational complexity (high-dimensional pixel arrays vs. structured text), lower feedback determinism (visual confirmation vs. binary exit codes), and substantially greater computational overhead~\citep{hong2024cogagent, zhang2025appagent}. Despite these trade-offs being unavoidable when automating the estimated 60--80\% of enterprise workflows locked behind graphical interfaces~\citep{van2018robotic}, GUI agents capture richer contextual signals---color-coded warnings, disabled buttons, spatial layout---that enable semantic error recovery impossible in text-only environments. As we demonstrate throughout this survey, RL provides the principled framework to navigate this complexity, transforming visual perception into robust task execution despite sparse rewards and dynamic interface evolution.

\paragraph{Scope and contributions.}
This survey provides the comprehensive analysis of RL techniques within the GUI agent domain, our contributions are:

\begin{itemize}
    \item \textbf{A principled RL taxonomy for GUI agents.} We categorize methods into three paradigms---offline RL, online RL, and hybrid strategies---and identify the dominant algorithmic families (DPO for offline RFT, GRPO for online/hybrid) together with their theoretical motivations and practical trade-offs.

    \item \textbf{Cross-cutting dimensional analysis.} We analyze three critical dimensions that cut across all paradigms: reward engineering (a three-tier pyramid from rule-based to LLM-as-Judge to learned rewards), data efficiency (world models, demonstration enhancement, self-improvement), and technical innovations in perception, memory, and multi-turn optimization.

    \item \textbf{Actionable insights and roadmap.} We distill three key findings---the reliability--scalability tension in reward design, the I/O latency wall driving world-model adoption, and the emergence of reasoning from structured action spaces---into a concrete research roadmap for the field, culminating in a broader perspective on digital inhabitants and the agent-native infrastructure they may ultimately require.
\end{itemize}

\section{Related Works}
\label{sec:related-works}
\paragraph{Surveys on RL for LLM alignment and reasoning.}
A growing body of work covers Reinforcement Learning for Large Language Models in text-centric domains---alignment via RLHF and DPO, reasoning enhancements for models like OpenAI o1 and DeepSeek-R1 that use MCTS and process rewards, and multi-agent RL~\citep{sun2024llm}. These surveys address static, text-only generation, which differs fundamentally from the multimodal, interactive setting of GUI agents where actions require precise visual grounding and real-time interface manipulation~\citep{cao2024survey, wang2024reinforcement, zhang2025landscape, zhang2025survey}.

\paragraph{Surveys on GUI agents.}
Recent reviews of autonomous GUI agents focus predominantly on architectural components (visual encoders, grounding modules) and supervised fine-tuning strategies, with some examining the trade-offs between API-based and GUI-based agents~\citep{zhang2025api}. While these works provide broad coverage of benchmarks and model architectures, they treat RL peripherally---lacking systematic analysis of reward engineering, exploration mechanisms, and the credit-assignment challenges unique to long-horizon GUI decision-making~\citep{nguyen2025gui, hu2025agents, liu2025llm, ning2025survey, sager2025comprehensive, tang2025survey, wang2024gui, zhang2024large}.

\paragraph{Gap and our contribution.}
\rev{Rather than positioning this paper as a general survey of GUI agents, we focus on the comparatively under-explored intersection of \textbf{Reinforcement Learning and GUI Agents}. Existing surveys provide valuable coverage of GUI, web, and broader computer-use agents, but they typically emphasize agent architectures, benchmarks, visual grounding, or supervised fine-tuning. In contrast, our work offers an early RL-centered synthesis of GUI and computer-use agents, with particular attention to training regimes, reward design, data efficiency, training infrastructure, and the design trade-offs induced by interactive GUI environments.} We differentiate our work along three axes:
\begin{itemize}
    \item We introduce a detailed taxonomy of RL paradigms adapted for GUI automation---offline, online, and hybrid---with cross-cutting analyses of reward engineering, data efficiency, and technical innovations.
    \item We conduct a systematic analysis of reward engineering techniques (rule-based, LLM-as-judge, learned) tailored to the sparse, delayed feedback characteristic of GUI tasks.
    \item We synthesize the rapid methodological advances of 2024--2026, offering a structured roadmap for this emerging field.
\end{itemize}

\section{Preliminaries}
\label{sec:preliminaries}

\subsection{Definition of GUI Agents}

GUI agents are intelligent systems capable of autonomously completing complex, cross-application tasks by perceiving on-screen information visually and simulating mouse and keyboard operations (such as clicking, typing, and scrolling), much like a human user.

\subsection{\texorpdfstring{The POMDP Formalism for GUI Agents}{The POMDP Formalism for GUI Agents}}

We formally model the interaction between the agent and the digital environment as a finite-horizon Partially Observable Markov Decision Process (POMDP). This framework is defined by the tuple:
\[
\mathcal{M} = (\mathcal{S}, \mathcal{A}, \rev{\mathcal{O},} \mathcal{T}, \rev{\mathcal{Z},} \mathcal{R}, \gamma, \rev{H})
\]

The latent state space $\mathcal{S}$ represents the underlying digital environment, including application memory, hidden DOM or operating-system state, network state, files, databases, and other variables that may not be directly visible to the agent. The observation space $\mathcal{O}$ contains the multimodal evidence available to the agent, typically high-resolution screenshots, OCR text, DOM snapshots, accessibility trees, or combinations of these signals. The observation model $\mathcal{Z}(o_t \mid s_t, a_{t-1})$ maps latent states and the previous action to the agent's current observation. The action space $\mathcal{A}$ consists of permissible low-level interface operations---including coordinate-based clicks ($x, y$), keyboard typing, scrolling, and drag-and-drop gestures---as well as higher-level tool or API actions in hybrid settings. The environment dynamics are governed by the transition function $\mathcal{T}(s_{t+1}\mid s_t,a_t)$, which is often stochastic due to network latency, dynamic page rendering, asynchronous UI updates, and hidden application state.

A critical challenge in this domain is the reward signal $\mathcal{R}: \mathcal{S} \times \mathcal{A} \rightarrow \mathbb{R}$, or an observation-based proxy when the latent state is not directly inspectable. In standard settings, rewards are sparse and binary ($+1$ for successful task completion, $0$ otherwise), necessitating the development of auxiliary dense reward functions to facilitate efficient learning. Because observations are partial, the interaction history is not part of the POMDP tuple but an input to the agent's decision process:
\[
\mathcal{H}_t = (o_0, a_0, \dots, o_{t-1}, a_{t-1}, o_t)
\]
which captures temporal dependencies essential for resolving ambiguity under partial observability. The objective of the GUI agent is to learn a history-conditioned policy $\pi(a_t \mid \mathcal{H}_t)$, or equivalently a belief-conditioned policy $\pi(a_t \mid b_t)$, that maximizes the expected cumulative discounted reward:
\[
J(\pi) = \mathbb{E}_{\pi}\left[\sum_{t=0}^{H} \gamma^t \mathcal{R}(s_t, a_t)\right]
\]
where $\gamma \in [0, 1]$ is the discount factor balancing immediate and future returns.

\subsection{Reinforcement Learning}

Reinforcement Learning (RL) is a computational framework for learning optimal decision-making strategies through interaction with an environment. Unlike supervised learning, which relies on static datasets of labeled examples, RL optimizes a policy $\pi$ by maximizing a cumulative reward signal gathered through trial-and-error exploration.

Within the GUI agent domain, RL is essential for solving \rev{partially observable}, long-horizon tasks where the optimal action depends on \rev{interaction} history and feedback is often sparse (i.e., received only upon task completion). RL algorithms are typically categorized by their usage of data (online vs.\ offline) and their optimization target (value-based vs.\ policy-based). Most state-of-the-art GUI agents employ policy gradient methods (e.g., PPO~\citep{schulman2017proximal}, GRPO~\citep{shao2024deepseekmath}) to fine-tune Multimodal Large Language Models (MLLMs), treating the model as a policy $\pi_\theta$ that maps visual observations \rev{and history/context representations} to executable actions.

\subsection{\texorpdfstring{\rev{Operational Definitions of Key Concepts}}{Operational Definitions of Key Concepts}}

\rev{We use the following operational meanings. A \textit{multimodal foundation model} or \textit{MLLM} refers to a pretrained model that conditions action prediction or reasoning on both language instructions and GUI observations, such as screenshots, OCR text, DOM snapshots, or accessibility trees; this definition does not assume that multimodality is always native rather than implemented through modular encoders. A \textit{GUI world model} denotes a learned model that predicts task-relevant interface dynamics, such as next observations, action outcomes, progress states, or failure modes. We reserve the term for models used to support planning, rollout generation, or policy learning, rather than for static datasets, handcrafted simulators, or simple rendering engines alone. \textit{System-2-style deliberation} describes observable slow-thinking behaviors in agents, including explicit multi-step planning, reflection, verification, search, or backtracking; we use this as an analogy to deliberative control, not as evidence that current agents possess human-like reasoning. Finally, \textit{verifiable reward} or \textit{environment feedback} refers to reward signals computed from observable task evidence, such as bounding-box matches, executable checks, URL or DOM changes, form states, or evaluator judgments. Such rewards are more auditable than free-form preference labels, but they are not assumed to be complete, ungameable, or reliable across all GUI tasks.}

\subsection{Background and Historical Evolution}

The trajectory of this field illustrates a shift from the broader concept of \textbf{Computer-Using Agents (CUAs)} to the specific paradigm of \textbf{GUI Agents}. While CUAs encompass any autonomous system operating a computer---including those relying on backend APIs, CLI commands, or hidden DOM states---GUI agents specifically target interaction through the visual frontend, mimicking human perception (vision) and actuator control (mouse/keyboard). This evolution reflects a transition from efficient but brittle backend automation to robust, general-purpose interaction that utilizes software exactly as humans do. We trace this development through three distinct phases.

\paragraph{Phase 1: Rule-based automation (1990s--2010s).}
Early UI automation relied on hard-coded scripts and Robotic Process Automation (RPA) tools that strictly replayed recorded sequences of coordinates or accessibility API calls. While effective for repetitive, static workflows, these systems were brittle: minor interface changes---such as a shifted button or a renamed text field---would cause catastrophic failure. They lacked perception and could not adapt to dynamic content~\citep{pasupat2018mapping}.

\paragraph{Phase 2: Deep reinforcement learning in isolated environments (2015--2022).}
With the rise of Deep Q-Networks (DQN)~\citep{mnih2013playing} and policy gradients, researchers began applying RL to GUI navigation. Early works like World of Bits~\citep{shi2017world} and MiniWoB++~\citep{liu2018miniwob} demonstrated that agents could learn to interact with web elements via trial and error. However, these agents were typically trained on small, synthetic environments using shallow neural networks. They struggled with generalization, often overfitting to specific DOM structures or visual layouts, and lacked the semantic understanding to process complex natural language instructions.

\paragraph{Phase 3: The multimodal LLM era (2023--present).}
The current breakthrough is driven by the integration of Multimodal Large Language Models (MLLMs) such as GPT-4V~\citep{openaiGPT4VisionTechnical}, Gemini~\citep{comanici2025gemini}, and Qwen-VL~\citep{wang2024qwen2}. These models provide agents with comprehensive semantic understanding of both screenshots (vision) and complex instructions (text). Unlike previous generations, modern GUI agents can reason about user intent, interpret novel interfaces zero-shot, and plan long-horizon tasks. Recent milestones include high-fidelity benchmarks like OSWorld~\citep{xie2024osworld} and WebArena~\citep{zhou2023webarena}, where agents now perform real-world tasks ranging from data entry to cross-application workflow management, though significant gaps in reliability and speed remain compared to human performance.

\subsection{Frontier Models}
\label{sec:frontier-models}

In this subsection, we provide an overview of state-of-the-art GUI agent models trained with RL-based methods, organized roughly chronologically along four major directions: closed-source commercial systems, open-source general-purpose agents, grounding-specialized models, and reasoning-enhanced architectures.

\definecolor{tax-edge}{RGB}{105,105,105}       

\definecolor{fnd-H}{RGB}{251,220,148}          
\definecolor{fnd-S}{RGB}{253,237,196}          
\definecolor{fnd-L}{RGB}{255,249,232}          
\definecolor{fnd-B}{RGB}{174, 98,  0}          

\definecolor{bnc-H}{RGB}{163,232,226}
\definecolor{bnc-L}{RGB}{218,248,245}
\definecolor{bnc-B}{RGB}{  0,112,100}

\definecolor{frt-H}{RGB}{190,234,193}
\definecolor{frt-S}{RGB}{215,243,218}
\definecolor{frt-L}{RGB}{234,249,235}
\definecolor{frt-B}{RGB}{ 34,120, 51}

\definecolor{rrl-H}{RGB}{183,216,255}
\definecolor{rrl-S}{RGB}{212,233,255}
\definecolor{rrl-L}{RGB}{232,244,255}
\definecolor{rrl-B}{RGB}{ 18, 83,195}

\definecolor{spc-H}{RGB}{218,183,255}
\definecolor{spc-S}{RGB}{235,212,255}
\definecolor{spc-L}{RGB}{246,233,255}
\definecolor{spc-B}{RGB}{ 92, 18,158}

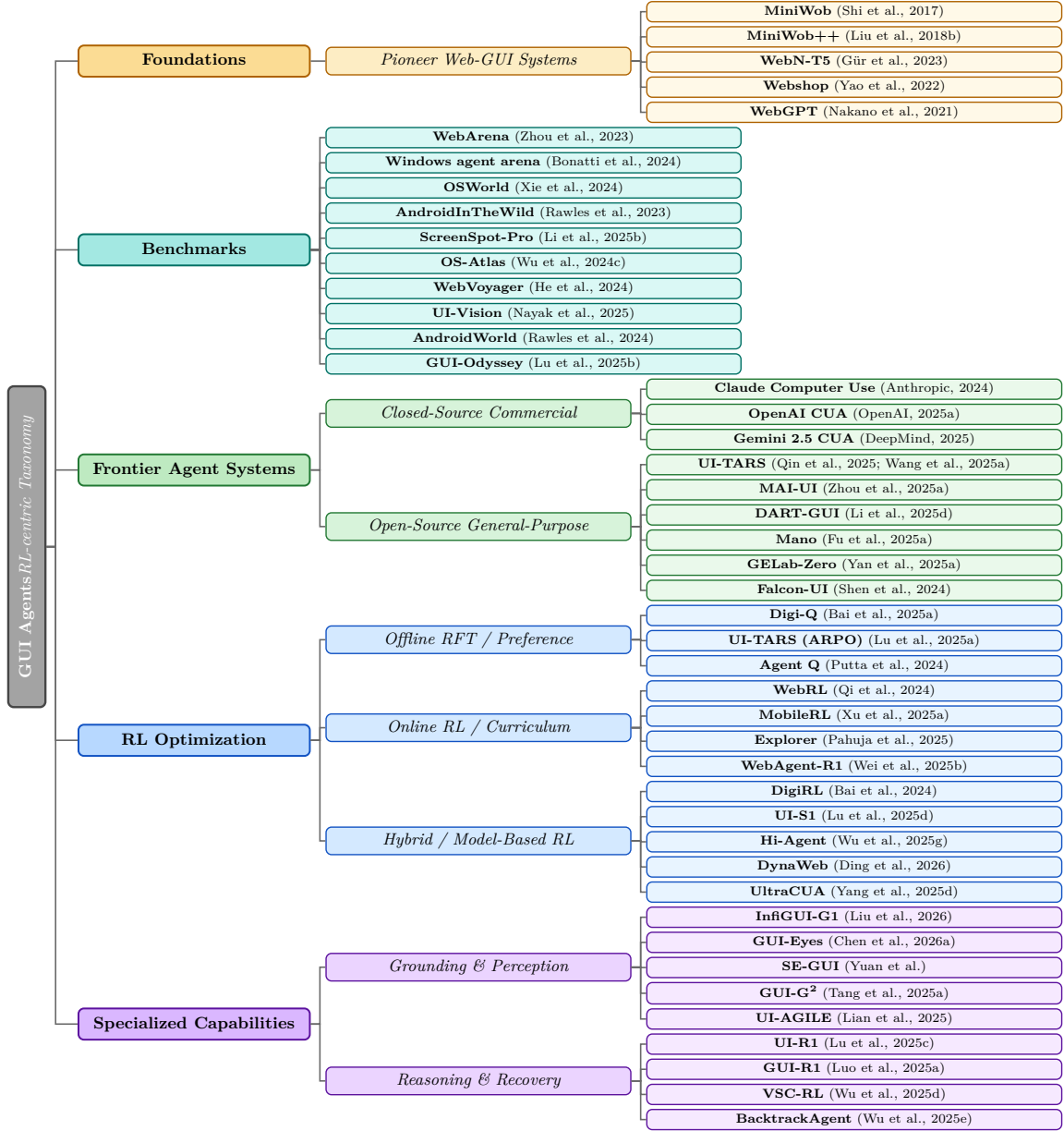
\begin{figure*}[t]
    \centering
    \resizebox{0.92\textwidth}{!}{%
        \begin{forest}
            forked edges,
            scale=0.92,
            transform shape,
            for tree={
                grow=east,
                reversed=true,
                anchor=base west,
                parent anchor=east,
                child anchor=west,
                base=center,
                font=\scriptsize,
                rectangle,
                draw=tax-edge,
                rounded corners=3pt,
                align=center,
                text centered,
                minimum width=6em,
                edge+={tax-edge, line width=0.8pt},
                s sep=2pt,
                l sep=8pt,
                inner xsep=3pt,
                inner ysep=1.5pt,
                line width=0.7pt,
                text width=6em,
                delay={where n children=0{font=\scriptsize}{}},
                ver/.style={rotate=90, child anchor=north, parent anchor=south, anchor=center},
            },
            [\textbf{GUI Agents}\newline{\normalsize\itshape RL-centric Taxonomy},
                ver, text width=18em,
                fill=gray!72, draw=gray!55!black,
                text=white, line width=1.2pt, font=\normalsize,
                inner xsep=5pt, inner ysep=5pt,
                [\textbf{Foundations},
                    fill=fnd-H, draw=fnd-B, text width=12em,
                    line width=1.0pt, font=\bfseries\small, inner ysep=3pt,
                    [\textit{Pioneer Web-GUI Systems},
                        fill=fnd-S, draw=fnd-B, text width=16em,
                        font=\itshape\small, inner ysep=2pt,
                        [\textbf{MiniWob}~\citep{shi2017world},
                            fill=fnd-L, draw=fnd-B,
                            text width=22em, align=left, font=\scriptsize]
                        [\textbf{MiniWob++}~\citep{liu2018miniwob},
                            fill=fnd-L, draw=fnd-B,
                            text width=22em, align=left, font=\small]
                        [\textbf{WebN-T5}~\citep{gur2023understanding},
                            fill=fnd-L, draw=fnd-B,
                            text width=22em, align=left, font=\small]
                        [\textbf{Webshop}~\citep{yao2022webshop},
                            fill=fnd-L, draw=fnd-B,
                            text width=22em, align=left, font=\small]
                        [\textbf{WebGPT}~\citep{nakano2021webgpt},
                            fill=fnd-L, draw=fnd-B,
                            text width=22em, align=left, font=\small]
                    ]
                ]
                [\textbf{Benchmarks},
                    fill=bnc-H, draw=bnc-B, text width=12em,
                    line width=1.0pt, font=\bfseries\small, inner ysep=3pt,
                    [\textbf{WebArena}~\citep{zhou2023webarena},
                        fill=bnc-L, draw=bnc-B,
                        text width=22em, align=left, font=\scriptsize]
                    [\textbf{Windows agent arena}~\citep{bonatti2024windows},
                        fill=bnc-L, draw=bnc-B,
                        text width=22em, align=left, font=\small]
                    [\textbf{OSWorld}~\citep{xie2024osworld},
                        fill=bnc-L, draw=bnc-B,
                        text width=22em, align=left, font=\small]
                    [\textbf{AndroidInTheWild}~\citep{rawles2023androidinthewild},
                        fill=bnc-L, draw=bnc-B,
                        text width=22em, align=left, font=\small]
                    [\textbf{ScreenSpot-Pro}~\citep{li2025screenspot},
                        fill=bnc-L, draw=bnc-B,
                        text width=22em, align=left, font=\small]
                    [\textbf{OS-Atlas}~\citep{wu2024atlas},
                        fill=bnc-L, draw=bnc-B,
                        text width=22em, align=left, font=\small]
                    [\textbf{WebVoyager}~\citep{he2024webvoyager},
                        fill=bnc-L, draw=bnc-B,
                        text width=22em, align=left, font=\small]
                    [\textbf{UI-Vision}~\citep{nayak2025ui},
                        fill=bnc-L, draw=bnc-B,
                        text width=22em, align=left, font=\small]
                    [\textbf{AndroidWorld}~\citep{rawles2024androidworld},
                        fill=bnc-L, draw=bnc-B,
                        text width=22em, align=left, font=\small]
                    [\textbf{GUI-Odyssey}~\citep{lu2025guiodyssey},
                        fill=bnc-L, draw=bnc-B,
                        text width=22em, align=left, font=\small]
                ]
                [\textbf{Frontier Agent Systems},
                    fill=frt-H, draw=frt-B, text width=12em,
                    line width=1.0pt, font=\bfseries\small, inner ysep=3pt,
                    [\textit{Closed-Source Commercial},
                        fill=frt-S, draw=frt-B, text width=16em,
                        font=\itshape\small, inner ysep=2pt,
                        [\textbf{Claude Computer Use}~\citep{anthropicIntroducingComputer},
                            fill=frt-L, draw=frt-B,
                            text width=22em, align=left, font=\scriptsize]
                        [\textbf{OpenAI CUA}~\citep{openaiComputerUsingAgent},
                            fill=frt-L, draw=frt-B,
                            text width=22em, align=left, font=\small]
                        [\textbf{Gemini 2.5 CUA}~\citep{geminiComputerUsingAgent},
                            fill=frt-L, draw=frt-B,
                            text width=22em, align=left, font=\small]
                    ]
                    [\textit{Open-Source General-Purpose},
                        fill=frt-S, draw=frt-B, text width=16em,
                        font=\itshape\small, inner ysep=2pt,
                        [\textbf{UI-TARS}~\citep{qin2025ui,wang2025ui},
                            fill=frt-L, draw=frt-B,
                            text width=22em, align=left, font=\small]
                        [\textbf{MAI-UI}~\citep{zhou2025mai},
                            fill=frt-L, draw=frt-B,
                            text width=22em, align=left, font=\small]
                        [\textbf{DART-GUI}~\citep{li2025efficient},
                            fill=frt-L, draw=frt-B,
                            text width=22em, align=left, font=\small]
                        [\textbf{Mano}~\citep{fu2025mano},
                            fill=frt-L, draw=frt-B,
                            text width=22em, align=left, font=\small]
                        [\textbf{GELab-Zero}~\citep{gelab_engine},
                            fill=frt-L, draw=frt-B,
                            text width=22em, align=left, font=\small]
                        [\textbf{Falcon-UI}~\citep{shen2024falcon},
                            fill=frt-L, draw=frt-B,
                            text width=22em, align=left, font=\small]
                    ]
                ]
                [\textbf{RL Optimization},
                    fill=rrl-H, draw=rrl-B, text width=12em,
                    line width=1.0pt, font=\bfseries\small, inner ysep=3pt,
                    [\textit{Offline RFT / Preference},
                        fill=rrl-S, draw=rrl-B, text width=16em,
                        font=\itshape\small, inner ysep=2pt,
                        [\textbf{Digi-Q}~\citep{bai2025digi},
                            fill=rrl-L, draw=rrl-B,
                            text width=22em, align=left, font=\scriptsize]
                        [\textbf{UI-TARS (ARPO)}~\citep{lu2025arpo},
                            fill=rrl-L, draw=rrl-B,
                            text width=22em, align=left, font=\small]
                        [\textbf{Agent Q}~\citep{putta2024agent},
                            fill=rrl-L, draw=rrl-B,
                            text width=22em, align=left, font=\small]
                    ]
                    [\textit{Online RL / Curriculum},
                        fill=rrl-S, draw=rrl-B, text width=16em,
                        font=\itshape\small, inner ysep=2pt,
                        [\textbf{WebRL}~\citep{qi2024webrl},
                            fill=rrl-L, draw=rrl-B,
                            text width=22em, align=left, font=\small]
                        [\textbf{MobileRL}~\citep{xu2025mobilerl},
                            fill=rrl-L, draw=rrl-B,
                            text width=22em, align=left, font=\small]
                        [\textbf{Explorer}~\citep{pahuja2025explorer},
                            fill=rrl-L, draw=rrl-B,
                            text width=22em, align=left, font=\small]
                        [\textbf{WebAgent-R1}~\citep{wei2025webagent},
                            fill=rrl-L, draw=rrl-B,
                            text width=22em, align=left, font=\small]
                    ]
                    [\textit{Hybrid / Model-Based RL},
                        fill=rrl-S, draw=rrl-B, text width=16em,
                        font=\itshape\small, inner ysep=2pt,
                        [\textbf{DigiRL}~\citep{bai2024digirl},
                            fill=rrl-L, draw=rrl-B,
                            text width=22em, align=left, font=\small]
                        [\textbf{UI-S1}~\citep{lu2025ui},
                            fill=rrl-L, draw=rrl-B,
                            text width=22em, align=left, font=\small]
                        [\textbf{Hi-Agent}~\citep{wu2025hi},
                            fill=rrl-L, draw=rrl-B,
                            text width=22em, align=left, font=\small]
                        [\textbf{DynaWeb}~\citep{ding2026dynaweb},
                            fill=rrl-L, draw=rrl-B,
                            text width=22em, align=left, font=\small]
                        [\textbf{UltraCUA}~\citep{yang2025ultracua},
                            fill=rrl-L, draw=rrl-B,
                            text width=22em, align=left, font=\small]
                    ]
                ]
                [\textbf{Specialized Capabilities},
                    fill=spc-H, draw=spc-B, text width=12em,
                    line width=1.0pt, font=\bfseries\small, inner ysep=3pt,
                    [\textit{Grounding \& Perception},
                        fill=spc-S, draw=spc-B, text width=16em,
                        font=\itshape\small, inner ysep=2pt,
                        [\textbf{InfiGUI-G1}~\citep{liu2026infiguig1},
                            fill=spc-L, draw=spc-B,
                            text width=22em, align=left, font=\scriptsize]
                        [\textbf{GUI-Eyes}~\citep{chen2026guieyes},
                            fill=spc-L, draw=spc-B,
                            text width=22em, align=left, font=\small]
                        [\textbf{SE-GUI}~\citep{yuan2025enhancing},
                            fill=spc-L, draw=spc-B,
                            text width=22em, align=left, font=\small]
                        [\textbf{GUI-G\textsuperscript{2}}~\citep{tang2025guig2},
                            fill=spc-L, draw=spc-B,
                            text width=22em, align=left, font=\small]
                        [\textbf{UI-AGILE}~\citep{lian2025ui},
                            fill=spc-L, draw=spc-B,
                            text width=22em, align=left, font=\small]
                    ]
                    [\textit{Reasoning \& Recovery},
                        fill=spc-S, draw=spc-B, text width=16em,
                        font=\itshape\small, inner ysep=2pt,
                        [\textbf{UI-R1}~\citep{lu2025uir1},
                            fill=spc-L, draw=spc-B,
                            text width=22em, align=left, font=\small]
                        [\textbf{GUI-R1}~\citep{luo2025guir1},
                            fill=spc-L, draw=spc-B,
                            text width=22em, align=left, font=\small]
                        [\textbf{VSC-RL}~\citep{wu2025vsc},
                            fill=spc-L, draw=spc-B,
                            text width=22em, align=left, font=\small]
                        [\textbf{BacktrackAgent}~\citep{wu2025backtrackagent},
                            fill=spc-L, draw=spc-B,
                            text width=22em, align=left, font=\small]
                    ]
                ]
            ]
        \end{forest}%
    }
    \caption{A taxonomy of representative GUI agent papers organised along five dimensions: \textbf{Foundations} (pioneer systems), \textbf{Environments \& Benchmarks} (evaluation platforms), \textbf{Frontier Systems} (deployed agents), \textbf{RL Optimization} (training methods), and \textbf{Specialized Capabilities} (perception, reasoning, and recovery).}
    \label{fig:gui_agent_taxonomy}
\end{figure*}

\paragraph{Closed-source commercial systems.}
Over the past year, RL has progressively expanded the frontier of GUI agents. \textbf{OpenAI's Computer-Using Agent (CUA)}~\citep{openaiComputerUsingAgent}, released in January 2025 alongside Operator, established the viability of RL-enhanced autonomous computer control, combining RLHF-style alignment with environment interaction RL on tasks with verifiable outcomes. \textbf{Anthropic's Claude Computer Use}~\citep{anthropicIntroducingComputer}, publicly available since late 2024 with continuous iterations, adopted a pure vision-based approach relying exclusively on screenshots, leveraging Constitutional AI~\citep{bai2022constitutional} combined with RLHF for safe yet effective tool usage. These closed-source systems demonstrated that multimodal foundation models could be successfully adapted for GUI automation through RL training.

\paragraph{Open-source general-purpose agents.}
Open-source efforts rapidly closed the gap with proprietary systems. The \textbf{UI-TARS} series from ByteDance Seed~\citep{qin2025ui,wang2025ui} represented a milestone, achieving 42.5\% on OSWorld through a two-stage ``SFT + RL'' paradigm (detailed in Section~\ref{sec:offline-rl}). \textbf{MAI-UI}~\citep{zhou2025mai} from Alibaba Tongyi Lab targeted mobile-first deployment with Dynamic RL Scaling across 512 parallel environments. \textbf{DART-GUI}~\citep{li2025efficient} addressed engineering challenges through decoupled multi-turn RL with adaptive data curation. \textbf{Mano}~\citep{fu2025mano} implemented a comprehensive three-stage hybrid pipeline (SFT $\rightarrow$ Offline RL $\rightarrow$ Online RL) using GRPO with composite rewards (Section~\ref{sec:hybrid-rl}). \textbf{GELab-Zero}~\citep{gelab_engine} from StepAI became the first GUI agent designed for edge deployment. Other notable contributions include \textbf{Falcon-UI}~\citep{shen2024falcon}, which pioneered ``understanding GUI before following instructions'' through large-scale unsupervised pretraining, and \textbf{DigiRL}~\citep{bai2024digirl}, which established baseline methodologies for offline-to-online transition (Section~\ref{sec:hybrid-representative}). A growing ecosystem of complementary agents has further expanded the frontier: Agent~S and Agent~S2~\citep{agashe2024agent,agashe2025agent} introduced agentic and generalist-specialist frameworks; CogAgent~\citep{hong2024cogagent} pioneered visual language models for GUI understanding; OS-Copilot~\citep{wu2024copilot} targeted generalist computer agents; UFO2~\citep{zhang2025ufo2} proposed a desktop AgentOS; and OpenCUA~\citep{wang2025opencua} provided open foundations for computer-using agents. Mobile-focused efforts include SpiritSight Agent~\citep{huang2025spiritsight}, Mobile-Agent-V3~\citep{ye2025mobile}, MagicGUI~\citep{tang2025magicgui}, and AgentCPM-GUI~\citep{zhang2025agentcpm}, while ShowUI~\citep{lin2025showui} and InfiGUIAgent~\citep{liu2025infiguiagent} advanced vision-language-action modeling. Further contributions include UITron~\citep{zeng2025uitron}, Ponder \& Press~\citep{wang2025ponder}, CoAct-1~\citep{song2025coact}, OmegaUse~\citep{zhang2026omegause}, ClickAgent~\citep{hoscilowicz2025clickagent}, Step-GUI~\citep{yan2025step}, and GTA1~\citep{yang2025gta1} for GUI test-time scaling.

\begin{figure}[h]
    \centering
    \includegraphics[width=0.85\textwidth]{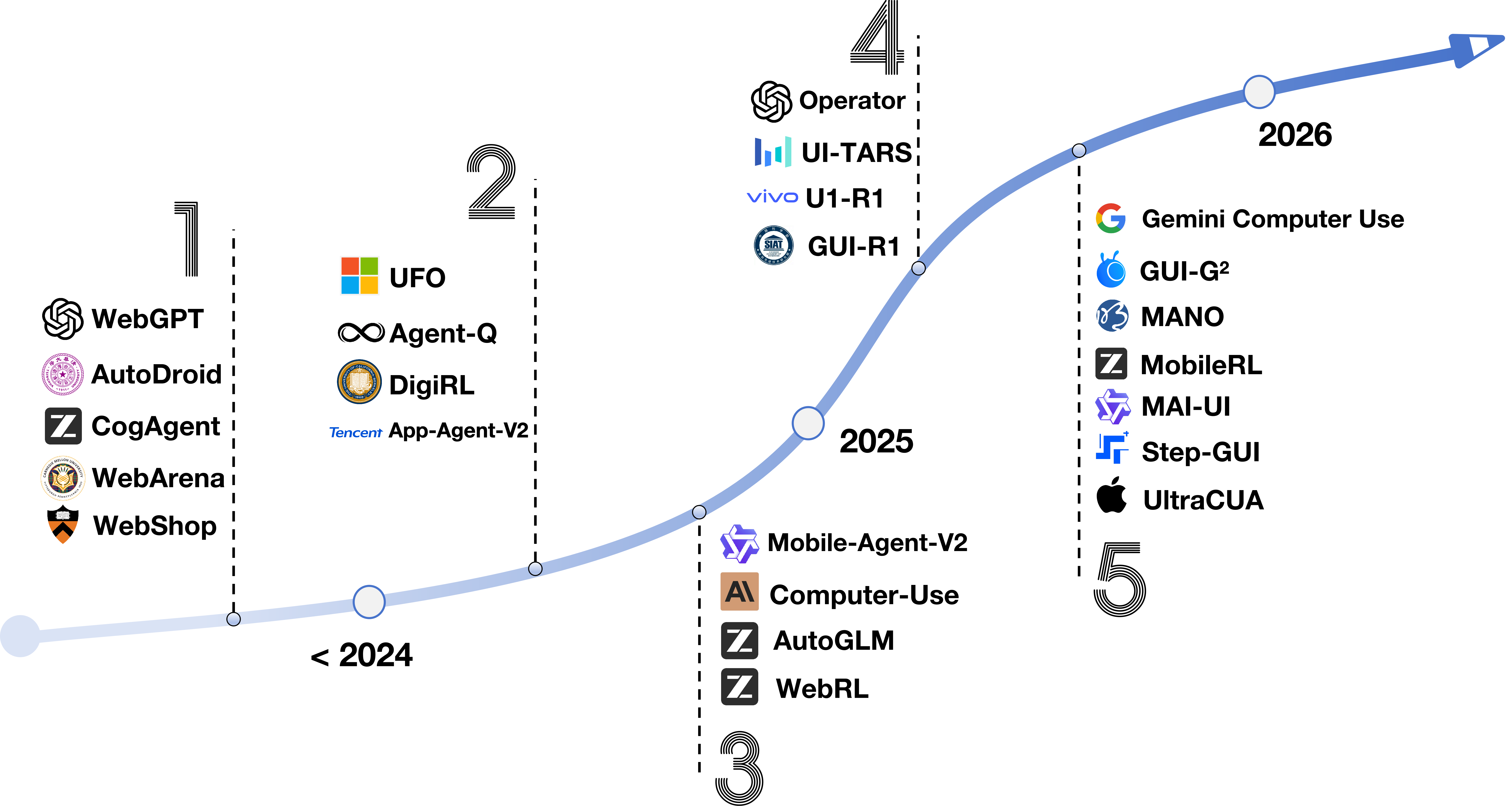}
    \caption{Timeline of GUI Agent Development.}
    \label{fig:timeline}
\end{figure}

\begin{table}[t]
\centering
\caption{Comparison of representative open-source GUI agent models with RL-based training. \textbf{Modality}: T=Text, I=Image, V=Video. \textbf{Platform}: D=Desktop, M=Mobile, W=Web.}
\label{tab:frontier_models}
\resizebox{\textwidth}{!}{%
\begin{tabular}{lcccccl}
\toprule
\textbf{Model} & \textbf{Release} & \textbf{Params} & \textbf{RL Algorithm} & \textbf{Modality} & \textbf{Platform} & \textbf{Key Innovation} \\
\midrule
UI-TARS-1.5/2 & 2025-04/09 & 2B--72B & SFT+RL & T/I & D/M/W & Thoughts-before-actions \\
MAI-UI & 2025-12 & 2B--235B & Dynamic RL Scaling & T/I/V & M & 512 parallel envs, edge-cloud \\
DART-GUI & 2025-09 & 7B & Decoupled RL & T/I & D/W & Adaptive data curation \\
Mano & 2025-09 & 7B & GRPO (3-stage) & T/I & W & Closed-loop data cycle \\
GELab-Zero & 2025-12 & 4B & Multi-turn RL & T/I & D/M & Edge deployment \\
InfiGUI-G1 & 2025-12 & 3B/7B & AEPO (RLVR) & T/I & D/M/W & Adaptive exploration reward \\
GUI-Eyes & 2026-01 & 7B & GRPO & T/I & D/M/W & Active perception \\
SE-GUI & 2025-05 & 7B & Self-evolutionary RL & T/I & D/M/W & Attention-based pseudo-labels \\
InfiGUI-R1 & 2025-04 & 3B & Actor2Reasoner RL & T/I & D/M & Error recovery rewards \\
UI-R1 & 2025-03 & 3B & Rule-based RL + DAST & T/I & D/M/W & Difficulty-adaptive thinking \\
GUI-R1 & 2025-04 & 7B & GRPO (RLVR) & T/I & D/M/W & Unified action space, data efficiency \\
UI-S1 & 2025-09 & 7B & Semi-online RL & T/I & M & Patch module, dual advantage \\
BacktrackAgent & 2025-05 & 7B & Verifier-guided RL & T/I & D/M/W & Error detection \& backtracking \\
VSC-RL & 2025-02 & 7B & Variational subgoal RL & T/I & M/W & SGC-ELBO objective \\
\bottomrule
\end{tabular}%
}
\end{table}

\paragraph{Grounding-specialized models.}
Visual grounding---precise mapping from natural language to screen coordinates---has emerged as a specialized focus for RL optimization, building on foundational work in universal visual grounding~\citep{gou2024navigating}, unified pure vision agents~\citep{xu2024aguvis, chen2026unify}, Aria-UI~\citep{yang2025aria}, and Phi-Ground~\citep{zhang2025phi} for perception, as well as visual test-time scaling for grounding~\citep{luo2025visual}. \textbf{InfiGUI-G1}~\citep{liu2026infiguig1} introduced AEPO (Adaptive Exploration Policy Optimization) to address insufficient exploration in continuous coordinate spaces (Section~\ref{sec:rule-based}). \textbf{GUI-Eyes}~\citep{chen2026guieyes} introduced active perception where models autonomously invoke visual tools before grounding (Section~\ref{sec:perception}). \textbf{SE-GUI}~\citep{yuan2025enhancing} proposed self-evolutionary RL leveraging attention maps as intermediate supervision. \textbf{GUI-G\textsuperscript{2}}~\citep{tang2025guig2} addressed the ``1-pixel deviation equals failure'' problem through Gaussian reward modeling (Section~\ref{sec:reward-eng}).

\paragraph{Reasoning-enhanced architectures.}
Reasoning capabilities have become increasingly central to GUI agent design. \textbf{InfiGUI-R1}~\citep{liu2025infigui} explicitly targeted transforming agents from ``Reactive Actors'' to ``Deliberative Reasoners'' through the Actor2Reasoner framework (AAAI 2026 Oral). \textbf{UI-R1}~\citep{lu2025uir1} adapted rule-based RL with DAST (Difficulty-Adaptive Slow-Thinking). \textbf{GUI-R1}~\citep{luo2025guir1} demonstrated extreme data efficiency through GRPO with verifiable rewards (detailed in Section~\ref{sec:offline-representative}). Semi-online and hybrid approaches have further bridged offline learning and online interaction: \textbf{UI-S1}~\citep{lu2025ui} introduced semi-online RL with trajectory patching (Section~\ref{sec:hybrid-representative}); \textbf{VSC-RL}~\citep{wu2025vsc} reconceptualized GUI control as subgoal-conditioned variational RL; and \textbf{BacktrackAgent}~\citep{wu2025backtrackagent} (EMNLP 2025) embraced error detection and recovery through a Generator-Verifier-Judger-Reflector architecture.

\paragraph{Foundation model backbones.}
Several multimodal foundation models serve as common backbones for GUI agent development. \textbf{Qwen3-VL}~\citep{bai2025qwen3vltechnicalreport} from Alibaba (2B--235B parameters) provides native Visual Agent capabilities with extended context (256K--1M tokens) and both Instruct and Thinking variants, serving as the backbone for MAI-UI, InfiGUI-G1, and GUI-Eyes. Other widely used backbones include Seed1.5-VL~\citep{guo2025seed1}, Kimi-VL~\citep{team2025kimi}, and InternVL3.5~\citep{wang2025internvl3}. \textbf{Seed-1.8}~\citep{bytedanceSeedNews} from ByteDance complements specialized GUI agents in a ``general brain + specialized executor'' architecture, with RL optimization on closed-loop business data. We provide a comprehensive timeline of GUI agent development in Figure~\ref{fig:timeline} and detailed information on open-source models in Table~\ref{tab:frontier_models}.

\section{RL Methods in GUI Agents}
\label{sec:rl-methods}

RL for GUI agents has branched into distinct methodological schools, each targeting a different bottleneck in the agent training lifecycle. We categorize the literature into three paradigms based on when and how the agent interacts with the environment. \textbf{Offline RL} focuses on learning from static datasets without environment interaction, enabling safe and scalable policy development. \textbf{Online RL} enables direct interaction with dynamic environments, optimizing policies through real-time trial and error. \textbf{Hybrid Strategies} bridge the gap between static pre-training and dynamic adaptation, including semi-online methods that simulate interaction dynamics on static data and staged training pipelines that combine offline initialization with online refinement. Across all paradigms, \textbf{Reinforcement Fine-Tuning (RFT)}---applying RL algorithms to fine-tune pretrained VLMs---serves as the dominant implementation approach, with offline methods typically employing DPO-based RFT and online methods employing PPO/GRPO-based RFT.

\paragraph{\rev{Taxonomy basis.}}
\rev{The offline/online/hybrid taxonomy in this survey is organized by \textit{interaction regime}: whether learning occurs from fixed trajectories, live environment rollouts, or a staged/semi-online combination of both. This axis is distinct from \textit{optimization algorithms} such as DPO, GRPO, CQL, IQL, or PPO, and from \textit{data sources} such as human demonstrations, synthetic trajectories, web logs, or live rollouts. A single GUI-agent method may therefore span multiple dimensions; for example, a hybrid pipeline may use human demonstrations for offline initialization, GRPO for policy optimization, and online rollouts for refinement. We choose interaction regime as the primary organizing axis because it directly determines data acquisition, exploration risk, reward availability, interaction cost, and deployment constraints in GUI environments. Algorithms and data sources are discussed within each regime rather than treated as co-equal taxonomy levels.}

\paragraph{\rev{Scenario-aware method selection.}}
\rev{The appropriate training and interaction strategy depends strongly on the deployment scenario. When a task exposes stable APIs, command-line tools, or structured backend interfaces, those channels are usually preferable because they provide lower latency, clearer state access, and more deterministic feedback; GUI interaction is most justified when such interfaces are unavailable, incomplete, or when visual layout, warnings, disabled controls, or cross-application context are themselves task-relevant. Within GUI-based settings, high-risk or irreversible workflows favor offline RL, preference optimization, and rule-based verification before any live deployment; resettable benchmarks, self-hosted web environments, and low-consequence mobile tasks are better suited to online RL because exploration and recovery data can be collected safely. Volatile, long-horizon, or expensive environments often call for hybrid pipelines: offline demonstrations or synthetic trajectories provide initialization, world models or simulators reduce rollout cost, and limited online refinement corrects distribution shift. Reward design follows the same scenario logic: precise grounding and form-completion tasks can use executable or rule-based rewards, while ambiguous user-intent tasks require layered evaluation with LLM-as-judge signals, human-audited subsets, and safety constraints.}


\subsection{Offline Reinforcement Learning}
\label{sec:offline-rl}

While online interaction yields the richest learning signal, the latency, cost, and safety risks of live GUI exploration have driven adoption of \textbf{Offline Reinforcement Learning}. This paradigm distills optimal behaviors from static datasets---web interaction logs, human demonstrations, synthesized trajectories---so that agents internalize complex reasoning patterns without incurring the expense or irreversibility of real-time trial-and-error. In GUI environments, where a single environment step can take 0.5--2\,s (network latency, rendering delays) and where erroneous actions may be irreversible (accidental deletions, unintended purchases), offline methods provide a critical pathway for safe, scalable agent development.

\subsubsection{Theoretical Foundations}

\paragraph{Core definition.} Offline RL (also called Batch RL) learns a policy $\pi(a|\mathcal{H})$ or $\pi(a|b)$ entirely from a fixed dataset $\mathcal{D} = \{(\mathcal{H}_i, a_i, r_i, \mathcal{H}'_i)\}_{i=1}^{|\mathcal{D}|}$ without any environment interaction during training. For GUI agents this constraint is not merely convenient but often necessary: real applications execute slowly, incur API costs, and risk irreversible side-effects.

\paragraph{Distribution shift and value overestimation.} The fundamental challenge in offline RL is \textit{distribution shift}. Standard Q-learning and its variants~\citep{van2016deep} update via $Q(h,a) \leftarrow r + \gamma \max_{a'} Q(h', a')$ for a history or belief representation $h$, which queries the value of actions that may never appear in the training dataset. When the learned policy selects an \textit{out-of-distribution (OOD) action} $a \notin \text{supp}(\pi_\beta)$, neural function approximators produce erroneously optimistic Q-values; the $\max$ operator then amplifies these errors, yielding a policy that performs well on paper but poorly in deployment. If $\pi_\beta$ denotes the behavior policy and $\pi$ the learned policy, the distributional mismatch $d^\pi(h,a) \neq d^{\pi_\beta}(h,a)$ causes errors to accumulate over long horizons.

\paragraph{Key technical approaches.} To address OOD action evaluation, the GUI agent community has adopted several principled strategies. \textbf{Conservative Q-Learning (CQL)}~\citep{kumar2020conservative} adds a regularization term to the Q-function loss that explicitly penalizes Q-values for OOD actions while rewarding Q-values for actions observed in the dataset, learning a \textit{lower bound} on the true Q-function that ensures policies remain conservative. \textbf{Implicit Q-Learning (IQL)}~\citep{kostrikov2021offline} avoids querying OOD actions entirely by using \textit{expectile regression} to estimate value functions within the support of the dataset, sidestepping the extrapolation problem altogether. \textbf{Decision Transformer (DT)}~\citep{chen2021decision} reframes RL as a \textit{sequence prediction problem}: given a sequence of states, actions, and a target return-to-go (RTG), the model autoregressively generates actions conditioned on achieving the specified return. By avoiding temporal-difference bootstrapping, DT circumvents value overestimation and naturally aligns with the Transformer architectures underlying modern VLMs.

\subsubsection{Offline RFT Methods}

Reinforcement Fine-Tuning (RFT) refers to applying RL algorithms to fine-tune a pretrained VLM that has already acquired basic instruction-following and GUI understanding capabilities through supervised fine-tuning (SFT). Unlike training RL from scratch, RFT leverages the strong prior knowledge embedded in pretrained models, dramatically accelerating convergence. The primary goals of RFT are to address hallucination problems that SFT cannot resolve and to improve credit assignment in multi-step reasoning tasks. Within the offline paradigm, two RFT approaches dominate.

\paragraph{Direct Preference Optimization (DPO).} DPO~\citep{rafailov2023direct} bypasses explicit reward modeling by exploiting the closed-form relationship between the optimal policy and the reward under a KL-constrained objective. Given preference pairs $(\tau^+, \tau^-)$ where $\tau^+$ is preferred, the loss is:
\[
\mathcal{L}_{\text{DPO}}(\theta) = -\mathbb{E}_{(\tau^+, \tau^-)} \left[ \log \sigma\!\left( \beta \log \frac{\pi_\theta(\tau^+)}{\pi_{\text{ref}}(\tau^+)} - \beta \log \frac{\pi_\theta(\tau^-)}{\pi_{\text{ref}}(\tau^-)} \right) \right]
\]
where $\pi_{\text{ref}}$ is the reference policy (typically the SFT checkpoint) and $\beta$ controls divergence. For GUI agents, preference pairs are constructed from successful versus failed trajectories in static datasets---sidestepping the instability of critic-network training that plagues PPO at VLM scale.

\paragraph{Offline GRPO with verifiable rewards.} Group Relative Policy Optimization (GRPO)~\citep{shao2024deepseekmath}, popularized by DeepSeek-R1, replaces the learned critic with group-relative advantage estimation. For a prompt $x$, GRPO samples a group $\{o_1, \ldots, o_G\}$ from the current policy and computes:
\[
\hat{A}_i = \frac{r(o_i) - \mu_{\mathbf{r}}}{\sigma_{\mathbf{r}}}, \quad \mu_{\mathbf{r}} = \frac{1}{G}\sum_{j=1}^G r(o_j), \quad \sigma_{\mathbf{r}} = \sqrt{\frac{1}{G}\sum_{j=1}^G (r(o_j) - \mu_{\mathbf{r}})^2}
\]
The policy update maximizes $\sum_i \hat{A}_i \log \pi_\theta(o_i|x)$ subject to a KL penalty against the reference policy. For GUI agents, $r(\cdot)$ is often a verifiable reward---coordinate-in-bounding-box checks, action-type matching, or format compliance---computable from static trajectory data without a learned reward model. By avoiding a learned value or critic model, GRPO can reduce memory relative to PPO-style actor-critic training. The magnitude of this saving, however, depends on model sharing, optimizer states, precision, sequence length, reference and reward models, and the distributed-training configuration.

\subsubsection{Representative Methods}
\label{sec:offline-representative}

We categorize representative offline RL and RFT methods for GUI agents into three principal technical approaches: value-based methods that learn action-value functions from static data, preference-based optimization that leverages trajectory comparisons, and policy gradient methods with verifiable rewards.

\paragraph{Value-based offline RL.} Value-based approaches train Q-functions to estimate long-term returns, enabling action selection through value maximization without requiring online interaction. \textbf{Digi-Q}~\citep{bai2025digi} exemplifies pure offline learning by first performing representation fine-tuning via SFT to ensure discriminative GUI state features, then training a lightweight MLP head on the \textit{frozen} VLM backbone using IQL or CQL variants. At inference, the policy employs \textit{Best-of-N re-ranking}: sampling $N$ candidate actions and selecting the highest-valued one, achieving 21.2\% improvement over prior offline methods on AndroidInTheWild~\citep{rawles2023androidinthewild}. This demonstrates that ``inference-time compute'' can effectively substitute for expensive online data collection. \textbf{DigiRL}~\citep{bai2024digirl} extends this paradigm through a two-stage offline-to-online framework; its offline stage uses the AndroidInTheWild dataset (715K trajectories) for initialization via filtered BC~\citep{torabi2018behavioral} or AWR~\citep{peng2019advantage} variants, while the online stage and full hybrid pipeline are detailed in Section~\ref{sec:hybrid-representative}.

\paragraph{Preference-based optimization.} DPO-based methods bypass explicit reward modeling by directly optimizing policies from trajectory preference pairs, offering training stability advantages for large VLMs. \textbf{UI-TARS}~\citep{qin2025ui} targets cross-platform automation (Android, Windows, Web) through native end-to-end screenshot processing, constructing preference pairs from successful versus failed trajectories. To address the sparse reward problem where entire batches may fail, UI-TARS introduces \textit{experience replay}~\citep{schaul2015prioritized}: maintaining a buffer of successful trajectories and sampling from it when current-batch rewards are uniformly zero, ensuring gradient validity. The ARPO~\citep{lu2025arpo} variant combining DPO with GRPO reached 20.4\% on OSWorld, substantially exceeding the 15.6\% SFT baseline. \textbf{Agent Q}~\citep{putta2024agent} advances preference optimization by integrating \textit{Monte Carlo Tree Search (MCTS)}~\citep{coulom2006efficient} for high-quality data synthesis. Guided MCTS simulates future web states using a value model for pruning, while a \textit{self-critique} mechanism enables AI-driven state evaluation during search. MCTS-discovered successful paths become DPO positive examples, with failed paths as negatives. This ``search is data'' philosophy improved Llama-3-70B's~\citep{dubey2024llama} zero-shot success on real-world web booking (e.g., OpenTable) from 18.6\% to 81.7\%---a 340\% relative gain---demonstrating that search can uncover complex trajectories inaccessible through random exploration.

\paragraph{Policy gradient with verifiable rewards.} GRPO-based methods, inspired by DeepSeek-R1's success in mathematical reasoning, optimize policies through group-relative advantage estimation with rule-based verifiable rewards. \textbf{GUI-R1}~\citep{luo2025guir1} and \textbf{UI-R1}~\citep{lu2025uir1} employ a \textit{unified action space} encoding clicks, swipes, and keyboard inputs, combined with meticulously designed binary rewards: for grounding tasks, $+1$ if the predicted coordinate falls within the target bounding box, $0$ otherwise; for multi-step tasks, sparse terminal rewards upon goal achievement (URL change, element match). GUI-R1 reported state-of-the-art results on ScreenSpot-Pro~\citep{li2025screenspot} and seven other benchmarks using 3K samples (GUI-R1-3K), approximately 0.02\% of OS-Atlas's~\citep{wu2024atlas} reported 13M training examples. Data-efficiency strategies are discussed further in Section~\ref{sec:data-eff}. Analysis revealed deliberation-like traces: models generated internal monologues (``first observe overall layout, then locate specific elements''), suggesting that rule-guided RLVR may reduce reliance on explicit reasoning annotations in sufficiently structured GUI tasks.

\subsubsection{Emerging Directions}
\label{sec:offline-emerging}

\paragraph{Visual-language alignment stability.} A critical challenge in applying RL to VLMs is \textit{visual forgetting}: aggressive parameter updates to optimize specific button-clicking behaviors may degrade the model's general visual recognition capabilities. Digi-Q's frozen-backbone strategy elegantly sidesteps this issue, but online RL methods require more sophisticated solutions such as KL-divergence constraints or Elastic Weight Consolidation (EWC)~\citep{kirkpatrick2017overcoming} to preserve visual grounding while optimizing action policies.

\paragraph{Toward System-2 GUI agents.} The success of reasoning-enhanced models like DeepSeek-R1 points toward GUI agents that are not merely reactive executors but deliberative reasoners. Reinforcing explicit reasoning processes through RL---where agents receive rewards for both correct actions \textit{and} valid reasoning chains---represents a promising frontier, with process reinforcement through implicit rewards~\citep{cui2025process} and AgentPRM~\citep{xi2025agentprm} offering principled approaches to step-level credit assignment. This direction is further discussed alongside the System~1/System~2 cognitive hybridization paradigm in Section~\ref{sec:hybrid-emerging} and Section~\ref{sec:future}.

\paragraph{Synthesis \& Insight: The Safety Barrier.} While Offline RL is often praised for its computational and data efficiency, its most critical role in GUI automation is acting as a \textbf{safety barrier}. Unconstrained online exploration by an untrained agent in a real operating system risks catastrophic and irreversible actions---deleting user data, sending unintended emails, or executing unauthorized financial transactions. By distilling the foundational semantics of UI interactions from static datasets, Offline RL confines the trial-and-error process to a safe proxy, ensuring ``common sense'' is acquired before the agent ever touches a live environment.

\subsection{Online Reinforcement Learning}
\label{sec:online-rl}

Online RL represents the most direct paradigm for GUI agent development, treating the GUI not as a static dataset but as a dynamic environment where agents refine policies through continuous trial. In contrast to offline RL, online RL enables agents to continuously interact with real environments, collect data in a streaming fashion, and update policies in real-time. This paradigm is particularly critical for GUI agents due to several domain-specific characteristics: \textit{environment volatility}, where software updates and UI redesigns introduce persistent distribution shifts that render static training obsolete; \textit{long-horizon sequential decision-making} with sparse terminal rewards, necessitating iterative trial-and-error to shape effective policies; \textit{generalization demands} across heterogeneous applications and platforms; and \textit{annotation bottlenecks}, where manual labeling of correct action sequences is expensive and fails to cover long-tail scenarios.

\subsubsection{Theoretical Foundations}

\paragraph{From imitation learning to online RL.} Early GUI agents relied on zero-shot prompting or Behavioral Cloning (BC)~\citep{florence2022implicit}, which treats action prediction as supervised classification. BC suffers from \textit{covariate shift}: a single erroneous prediction at step $t$ pushes the agent into observations and histories absent from the training distribution. Under a per-step imitation error $\epsilon$, the expected number of mistakes scales as $\mathcal{O}(H\epsilon)$ over horizon $H$, while worst-case cumulative cost degradation can scale as $\mathcal{O}(H^2\epsilon)$~\citep{ross2011reduction}. This motivates online RL, where the agent can collect recovery experience from its own mistakes.

\paragraph{POMDP formulation.} Online RL for GUI agents is formalized as a Partially Observable Markov Decision Process (POMDP). Unlike offline methods, online agents actively interact with environments, enabling exploration and correction to learn recovery from erroneous states, as well as non-stationarity adaptation to handle dynamic GUI changes (e.g., loading speeds, popups).

\paragraph{Observation and action space heterogeneity.} GUI observations are highly heterogeneous, combining visual (pixels) and structural (DOM/accessibility) modalities while only partially exposing the latent application state. The action space is similarly mixed, featuring both discrete types (Click, Type) and continuous parameters. Since traditional algorithms like DQN struggle with hybrid spaces, mainstream approaches favor policy gradient methods (e.g., PPO, GRPO) with specialized action decoding.

\paragraph{Sparse rewards and reward engineering.} Acquiring reward signals is a major bottleneck. Task success (e.g., a multi-step purchase) is extremely sparse, delayed, and difficult to verify automatically. Consequently, designing dense reward functions or using Model-as-a-Judge evaluators (see Section~\ref{sec:reward-eng}) has become essential.

\paragraph{Sample efficiency and interaction cost.} Online RL is bounded by the slow speed of real-world environment interactions (e.g., rendering and network latency). Training requires extensive sample collection, making techniques like curriculum learning, offline pre-training, and semi-online methods critical for improving sample efficiency.

\subsubsection{Representative Methods}
\label{sec:online-representative}

\paragraph{Curriculum-based online learning.} Curriculum learning addresses the cold-start problem where random exploration yields insufficient positive rewards. WebRL ~\citep{qi2024webrl} introduces a \textit{Self-Evolving Online Curriculum} comprising task generation using teacher models and a \textit{Failure Set Strategy} that collects unsuccessful tasks and generates simplified variants, ensuring training tasks remain within the agent's ``Zone of Proximal Development.'' Related approaches include Curriculum-RLAIF~\citep{li2025curriculum}, which combines curriculum strategies with AI feedback, and RLAIF~\citep{lee2023rlaif}, which demonstrated the viability of AI-generated feedback as an alternative to human feedback. WebRL additionally trains an \textit{Outcome-Supervised Reward Model (ORM)}~\citep{yu2024ovm} that judges trajectory success from final states, providing stronger generalization than rule-based checkers. KL-divergence constraints and experience replay filtering prevent policy drift while preserving general capabilities.

\paragraph{Difficulty-adaptive policy optimization.} Mobile GUI environments exhibit heavy-tailed task difficulty distributions where standard RL algorithms are dominated by simple task gradients. MobileRL ~\citep{xu2025mobilerl} addresses this through \textit{AdaGRPO}, introducing difficulty weighting based on historical success rates---lower success rates yield higher gradient weights. \textit{Shortest-Path Reward Adjustment (SPA)} suppresses reward hacking by penalizing redundant operations, while \textit{Failure Curriculum Filtering (FCF)} temporarily removes tasks with persistent zero success rates.

\paragraph{Offline-to-online transition frameworks.} Tabula rasa online RL is impractical due to low exploration efficiency and potentially dangerous operations. DigiRL ~\citep{bai2024digirl} proposes a canonical two-stage paradigm combining offline initialization with online fine-tuning, pioneering VLMs as automatic evaluators for task completion judgment. We discuss DigiRL's full hybrid pipeline---including the Digi-Q algorithm and Best-of-N inference---in Section~\ref{sec:hybrid-representative}.

\paragraph{End-to-end multi-turn optimization.} WebAgent-R1 ~\citep{wei2025webagent} proposes \textit{Multi-Turn GRPO (M-GRPO)}, treating entire interaction trajectories as optimization samples rather than single-step actions, enabling agents to learn ``delayed gratification.'' Recent work has extended multi-turn RL optimization further: Sweet-RL~\citep{zhou2025sweet} introduced reward strategies specifically designed for multi-turn LLM agents, while RLTHF~\citep{xu2025rlthf} proposed targeted human feedback mechanisms for fine-grained turn-level credit assignment. \textbf{HGPO}~\citep{he2026hierarchy} further sharpens this line by addressing historical-context inconsistency in stepwise GRPO/GiGPO-style updates: when prompts depend on long interaction histories, optimizing each step against a stale or partially reconstructed context can assign credit to actions under a different state than the one seen at execution time. This issue is especially acute for GUI agents whose prompts interleave screenshots, summaries, tool traces, and memory snippets. \textit{Dynamic Context Compression} addresses context window explosion by having agents output observation summaries at each step, while \textit{Parallel Trajectory Rollout} improves training diversity.

\paragraph{Grounding-specialized methods.} InfiGUI-G1~\citep{liu2026infiguig1} discovers that for concrete grounding tasks, forcing Chain-of-Thought~\citep{wei2022chain} reasoning actually \textit{decreases} precision due to hallucinations. \textit{Fast Thinking Templates} suppress reasoning and directly regress coordinates, with ``System 1'' mode outperforming ``System 2'' for grounding. UI-AGILE ~\citep{lian2025ui} employs continuous distance-based rewards $R = \max(0, 1 - \text{distance}/\text{threshold})$ providing denser gradients than binary hit/miss rewards, combined with \textit{Cropping-Based Resampling} for small element recognition.

\paragraph{Exploration-driven data synthesis.} Explorer ~\citep{pahuja2025explorer} addresses cold-start through a multi-agent pipeline where ``explorers'' randomly walk through environments discovering novel states, and ``annotators'' reverse-engineer natural language instructions reaching these states, synthesizing high-quality trajectories for subsequent online training.

\paragraph{Infrastructure-aware online RL.} Beyond algorithmic exploration, recent systems emphasize that GUI RL quality depends on rollout infrastructure and signal hygiene. \textbf{AgentCPM-Explore}~\citep{chen2026agentcpm} is representative: it studies RL under noisy real I/O, combines reward-signal denoising with context compression, and treats trajectory collection as a systems problem rather than a purely policy-optimization problem. This makes it a useful concrete anchor for the I/O-wall argument developed later in Section~\ref{sec:data-eff}.

\subsubsection{Emerging Directions}
\label{sec:online-emerging}

\paragraph{Curriculum learning as the dominant paradigm.} From WebRL's ``self-evolving failure sets'' to MobileRL's ``failure curriculum filtering,'' all high-performance frameworks abandon random sampling in favor of curriculum learning variants. This reflects the enormous heterogeneity in GUI task spaces---agents must actively select training data appropriate to current capabilities for efficient learning. Future agents will function not merely as learners but as ``self-educators'' capable of designing their own practice problems.

\paragraph{Separation of reasoning and execution.} WebRL and MobileRL emphasize reasoning for planning, while InfiGUI-G1 demonstrates that reasoning interferes with precise grounding. This suggests future architectures will adopt dual-system designs---a theme we elaborate in Section~\ref{sec:hybrid-emerging} and Section~\ref{sec:future}.

\paragraph{Model-as-a-judge normalization.} As environment complexity increases, writing rule-based reward functions becomes impractical. DigiRL's VLM evaluator and WebRL's ORM mark the arrival of the AI-evaluating-AI era. Reward engineering is transforming into reward modeling (Section~\ref{sec:llm-judge}). While this solves sparse reward problems, it introduces new risks---\textit{reward hacking}---and ensuring evaluation model robustness represents the next research frontier.

\paragraph{Synthesis \& Insight: The I/O Wall.} The fundamental bottleneck in scaling Online RL for GUI agents lies not in algorithmic maturity, but in \textbf{The I/O Wall}. Unlike the microsecond transitions in simulated games like Atari or Go, real-world GUI interactions suffer from severe latency---network requests, page rendering, DOM parsing, and UI animations easily push single-step processing times to 0.5--2.0 seconds. This staggering environment feedback latency structurally caps sample collection throughput. Consequently, the core contradiction in Online RL is resolving this I/O gap, shifting the focus from purely algorithmic improvements to system-level innovations like heavily optimized simulators, parallelized cloud browser rendering, or asynchronous multi-agent rollouts.

\subsection{Hybrid Strategies}
\label{sec:hybrid-rl}

Pure offline RL suffers from distribution shift---where agents fail to recover from states unseen during training---while pure online RL is sample-inefficient and carries substantial risks in real operating system environments. Hybrid strategies attempt to bridge this gap through complementary approaches that combine the safety of offline methods with the adaptability of online exploration. These approaches have emerged as the dominant paradigm for training state-of-the-art GUI agents, achieving performance levels that neither pure offline nor pure online methods can match independently.

\subsubsection{Theoretical Foundations}

\paragraph{Core motivation.} Hybrid strategies resolve the tension between offline and online RL: offline methods provide safe policy initialization but suffer from distribution shift, while online methods enable adaptive exploration but are sample-inefficient and risky. Hybrid approaches leverage complementary advantages by using offline data for warm starts, online/semi-online rollouts for distribution correction, hierarchical architectures for long-term credit assignment, and world models for low-cost latent exploration.

\paragraph{Hybrid optimization objective.} Hybrid RL extends the traditional discounted return $J(\pi)$ with auxiliary losses:
\[
L_{\text{hybrid}} = \lambda_1 L_{\text{RL}} + \lambda_2 L_{\text{BC}} + \lambda_3 L_{\text{Reasoning}}
\]
Here, $L_{\text{RL}}$ optimizes long-term returns (e.g., PPO/GRPO), $L_{\text{BC}}$ prevents catastrophic forgetting early in training via behavioral cloning, and $L_{\text{Reasoning}}$ enforces logical planning via chain-of-thought generation. Loss weights typically shift from $L_{\text{BC}}$ to $L_{\text{RL}}$ as training progresses.

\paragraph{Hybrid action space formulation.} Advanced agents integrate distinct action modalities: \textit{atomic actions} ($\mathcal{A}_{\text{low}}$), which are universal but inefficient pixel-level primitives (e.g., \texttt{click(x,y)}), and \textit{semantic/tool actions} ($\mathcal{A}_{\text{high}}$), which are efficient but environment-dependent API operations (e.g., \texttt{checkout()})~\citep{song2025beyond}. Agents dynamically route between these action spaces to balance visual robustness with execution efficiency.

\paragraph{Key technical approaches.} The community has developed several core strategies. \textit{Semi-online learning} simulates online dynamics (e.g., trajectory patching) on offline datasets, allowing models to learn error recovery without live interaction costs. \textit{Staged training pipelines} sequentially transition from offline initialization to targeted online exploration, minimizing catastrophic failure risks in real environments. \textit{Hierarchical RL (HRL)} decomposes tasks into a Planner (generating semantic subgoals) and an Executor (performing UI actions), isolating complexity and mitigating sparse rewards. \textit{Model-Based RL (MBRL)} employs UI world models to dream realistic state transitions~\citep{gu2024your}, massively accelerating exploration in latent space before grounding with limited online samples.

\subsubsection{Representative Methods}
\label{sec:hybrid-representative}

We highlight six principal technical approaches within hybrid RL methods, each addressing specific bottlenecks in GUI agent training. In semi-online reinforcement learning, \textbf{UI-S1}~\citep{lu2025ui} simulates online dynamics on static data using a \textit{Patch Module} to correct out-of-distribution actions and a dual-level advantage function, maintaining offline throughput while mimicking online error-correction. For offline-to-online transition, \textbf{DigiRL}~\citep{bai2024digirl} employs a two-stage pipeline---offline initialization followed by targeted online fine-tuning---using \textit{Digi-Q} with a frozen VLM backbone and \textit{Best-of-N} sampling to reach a 67.2\% success rate on AitW. To handle long-horizon tasks, \textbf{Hi-Agent}~\citep{wu2025hi} introduces hierarchical planning and execution, jointly training a semantic \textit{Planner} and a UI \textit{Executor} via GRPO and a \textit{Foresight Advantage Function}, unlocking 87.9\% on AitW. Related hierarchical approaches include HiPER~\citep{peng2026hiper}, which addresses credit assignment through hierarchical RL with structured reward decomposition, probabilistic subgoal representations for HRL~\citep{wang2024probabilistic}, and \textbf{MiRA}~\citep{wang2026subgoal}, which operationalizes milestone-based planning and potential shaping for web navigation. MiRA is particularly relevant because it turns the otherwise abstract idea of subgoal-driven GUI RL into concrete intermediate objectives, grounding multi-tier reward design in measurable navigation progress. Addressing latency and safety, \textbf{DynaWeb}~\citep{ding2026dynaweb} leverages world model-augmented learning by efficiently ``dreaming'' trajectory rollouts within a \textit{Web World Model (WWM)}. In the domain of hybrid action spaces, \textbf{UltraCUA}~\citep{yang2025ultracua} unifies visual primitives and programmatic API tools, training the agent to flexibly route between universal visual fallbacks and fast deterministic executions. Finally, \textbf{UI-AGILE}~\citep{lian2025ui} demonstrates training-inference dual enhancement by combining dense IoU-based grounding rewards with grid-based partitioned reasoning at inference, improving ScreenSpot accuracy by 23\%.

\subsubsection{Emerging Directions}
\label{sec:hybrid-emerging}

\paragraph{Unified cross-ecosystem agents.} Current agents typically specialize in single platforms. However, real user workflows are cross-ecosystem (capturing images on mobile, editing on desktop, emailing via web). Future hybrid architectures must enable automatic alignment of interaction logic across heterogeneous operating systems through RL, achieving ``train once, deploy everywhere'' generalization (see Section~\ref{sec:future}).

\paragraph{Continual learning and sim-to-real transfer.} World model-based approaches like DynaWeb face simulation-reality gaps---real web environments contain CAPTCHAs, dynamic advertisements, and network failures that are difficult for world models to perfectly predict. Future hybrid strategies must incorporate domain randomization during dreaming and online adaptation during deployment, enabling agents to leverage test-time feedback for continuous policy refinement without catastrophic forgetting.

\paragraph{Privacy-aware hybrid learning.} As agents access sensitive data, future hybrid strategies must integrate \textit{Privacy Critics} that impose penalties for high-risk operations, implementing Safe RL principles within the optimization framework. Constrained RL formulations~\citep{zhang2024constrained} and worst-case optimization~\citep{yang2021wcsac} provide theoretical foundations for enforcing safety constraints during policy optimization (see also Section~\ref{sec:future}).

\paragraph{Efficient edge deployment.} Hybrid strategies involving world models and chain-of-thought reasoning substantially increase inference computation. Deploying 7B--30B parameter models on mobile devices faces energy and latency challenges. Future research directions include distillation and quantization: using powerful hybrid RL agents as teachers to guide lightweight student models (1B--3B parameters) that bypass heavy reasoning processes and directly learn optimal action mappings for efficient on-device execution.

\paragraph{Synthesis \& Insight: Cognitive Stratification.} Ultimately, Hybrid strategies embody a profound structural shift toward \textbf{Cognitive Stratification}. Rather than viewing offline and online phases as mere technical prerequisites for training efficiency, they serve distinct cognitive purposes in the agent's evolution. The offline phase initializes the agent's ``System 1''---the fast, intuitive ``common sense'' required to robustly perceive components, interpret icons, and execute safe atomic actions. Built upon this foundation, the online phase acts as the crucible for ``System 2''---honing the agent's intuition, long-horizon planning, error recovery, and complex reasoning in dynamic novel environments. This layered evolution gracefully resolves the tension between execution efficiency and goal robustness.

\paragraph{Cognitive hybridization: System 1 and System 2.} A recurring theme across all paradigms (Sections~\ref{sec:offline-emerging} and~\ref{sec:online-emerging}): inspired by dual-process theory, future GUI agents will likely integrate fast, intuitive ``System 1'' modules for routine operations with slow, deliberative ``System 2'' modules for complex reasoning. RL can optimize the routing between these cognitive modes---learning when quick reactions suffice versus when deep thinking is required. Novel policy optimization approaches such as group-in-group optimization~\citep{feng2025group}, HGPO~\citep{he2026hierarchy}, and multi-agent RL with state modelling~\citep{kontogiannis2025enhancing} offer complementary perspectives on structuring agent interactions and optimization groups. \textbf{ERL}~\citep{shi2026experiential} adds a complementary mechanism: structured reflection can transform sparse terminal feedback into learnable intermediate signals and consolidate successful revisions across attempts. Evidence from GUI-R1's emergent reasoning (Section~\ref{sec:offline-representative}) and InfiGUI-G1's finding that explicit CoT \emph{hurts} grounding (Section~\ref{sec:perception}) suggests that cognitive hybridization represents a promising frontier.

\section{Key Dimensions}
\label{sec:key-dimensions}

Building upon the method-centric overview of RL paradigms presented in Section~\ref{sec:rl-methods}, this section adopts a \emph{dimension-centric} perspective. Specifically, it analyzes three critical, cross-cutting dimensions that govern the design of RL-based GUI agents across all paradigms: \emph{reward engineering}, \emph{data efficiency}, and \emph{technical innovations} (encompassing algorithmic, perceptual, and memory-related advances). Each dimension addresses foundational challenges in GUI automation that parallel, yet remain distinct from, the difficulties encountered in traditional RL domains. To highlight overarching design principles and facilitate cross-method comparisons, the following discussion draws upon concrete instantiations introduced in Section~\ref{sec:rl-methods}, avoiding redundant descriptions of individual systems.

\subsection{Reward Engineering}
\label{sec:reward-eng}

Unlike classical RL environments that provide explicit rewards, GUI automation requires interpreting complex visual and semantic evidence to assess task completion~\citep{nguyen2025gui}. Formally, the ideal GUI reward combines a terminal indicator and a dense shaping function $\phi(s_t, a_t, g)$:
\[
\mathcal{R}^*(s_t, a_t) = \underbrace{\mathbb{1}[\text{task completed at } t]}_{\text{terminal}} + \underbrace{\lambda \cdot \phi(s_t, a_t, g)}_{\text{dense shaping}}
\]
Designing $\phi$ is notoriously difficult: sparse signals hinder learning, while overly dense ones provoke reward hacking. To navigate this accuracy--generality trade-off, current literature converges on a three-tier taxonomy: \emph{rule-based} rewards exploiting UI structures, \emph{LLM-as-judge} rewards evaluating via foundation models, and \emph{learned} rewards parameterized and optimized alongside the policy. 

\rev{\textbf{The shift toward verifiable environment feedback.} A critical overarching insight is that reward engineering is shifting from ``manually defined formatting heuristics'' toward \textbf{environment-feedback verification}. Many GUI environments afford relatively objective task evidence---such as URL transitions, DOM alterations, form states, or observable database changes---which makes them well suited to \textbf{Reinforcement Learning with Verifiable Rewards (RLVR)} when the task objective can be operationalized. State-of-the-art systems increasingly anchor their optimization on these auditable environmental signals, but such signals remain incomplete for ambiguous goals, user-specific preferences, and safety constraints. \textbf{Information-aware credit assignment (ICA)}~\citep{pang2026ica} strengthens this argument by explicitly tying credit to informative observations rather than treating every historical token or UI state as equally relevant. For GUI and web agents, this reinforces the case for visual-first observations: screenshots, highlighted regions, and verifiable state changes often carry denser credit information than brittle HTML parser traces alone.}

\begin{figure}[h]
    \centering
    \includegraphics[width=0.95\textwidth]{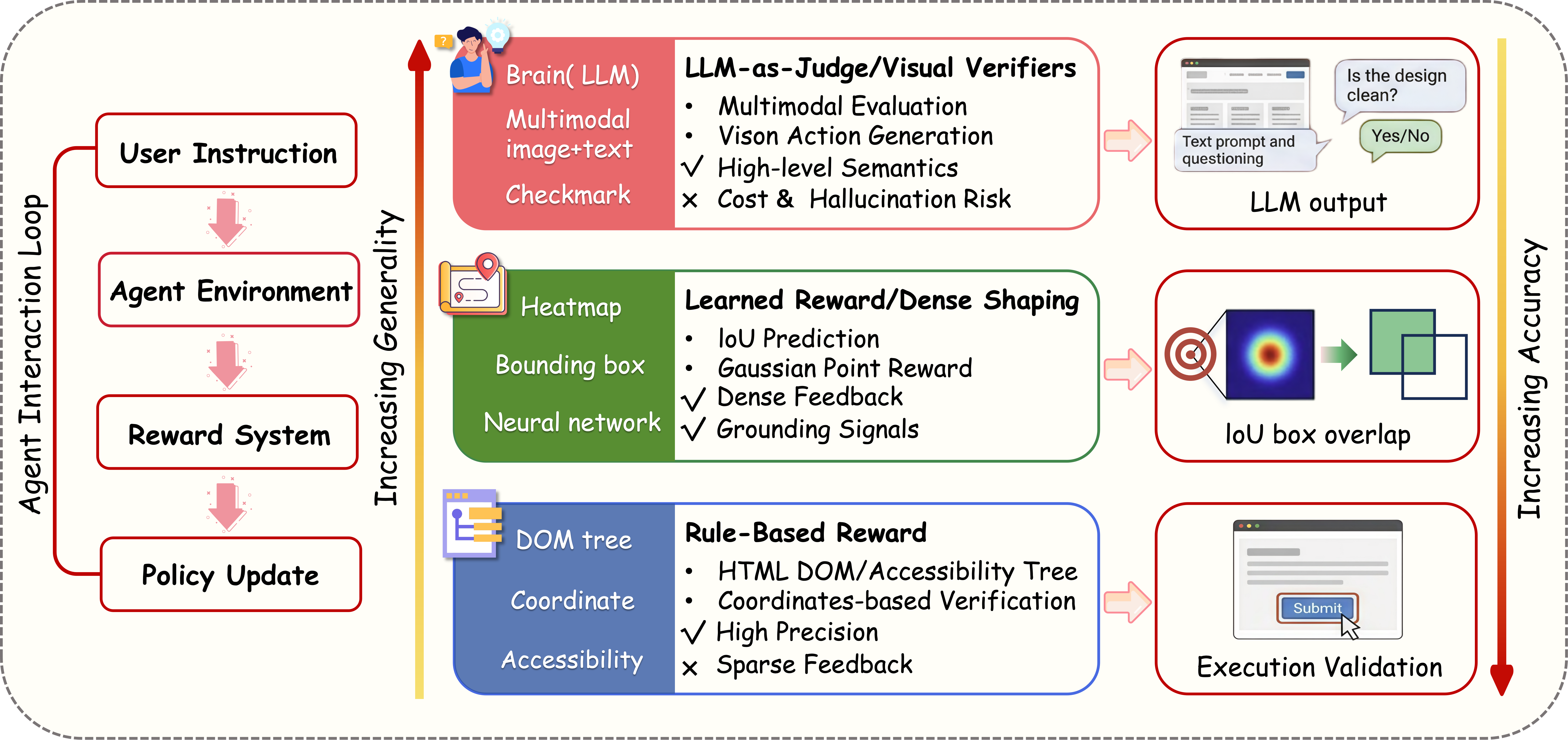}
    \caption{The Reward Engineering Pyramid balances accuracy and generality for GUI Agents: rule-based rewards (base) offer precision; learned rewards (middle) provide dense signals; LLM-as-Judge (apex) enables broad semantic task handling with hallucination risks.}
    \label{fig:reward-pyramid}
\end{figure}

\subsubsection{Rule-Based Rewards}
\label{sec:rule-based}

\rev{Rule-based rewards offer interpretable and computationally cheap signals by leveraging structured OS metadata (e.g., DOM trees, bounding boxes) to build explicit scoring functions without learned models. A pivotal question arises: \emph{why do seemingly rigid, rule-based systems remain useful optimization signals for SOTA models?} The practical answer lies in their provision of \textbf{auditable feedback}. In an era where LLM policies can exploit semantic loopholes or subjective evaluations of AI judges (reward hacking), rule-based verifiable rewards provide narrower but more reproducible task evidence. They anchor the feedback loop in observable interface changes or annotated targets, encouraging execution correctness rather than merely plausible-looking actions. Their design space spans from binary outcomes to dense continuous shaping, but they should be treated as task-specific approximations rather than complete or ungameable ground truths.}

\paragraph{From binary outcomes to continuous shaping.}
Binary rewards (success\,${}={+1}$, failure\,${}={0}$) present clean targets but suffer from credit-assignment issues. \textbf{UI-R1} \citep{lu2025uir1} mitigates this by decomposing rewards into verifiable action and format checks. Such \emph{verifiable} rewards---where ground truth is algorithmically determined---enable GRPO-style optimization without explicit reward models, allowing agents to consistently outperform heavily supervised baselines. Similarly, \textbf{BTL-UI}~\citep{zhang2025btl} employs a composite verifiable reward ($R = R_{\text{format}} + R_{\text{blink}} + R_{\text{link}}$) simulating a human ``Blink-Think-Link'' process. By integrating format compliance, region IoU, and action matching, this richer decomposition significantly boosts task success rates.

A fundamental limitation of binary rewards is that a prediction one pixel outside the bounding box is penalized identically to one that is entirely off-screen, producing vanishing gradients for all but the most accurate samples. Two approaches address this by introducing continuous spatial shaping. \textbf{GUI-G$^{2}$} \citep{tang2025guig2} replaces the binary indicator with a \emph{Gaussian point reward} centered on the element centroid, with variance proportional to the bounding-box area, and adds a complementary \emph{coverage reward} measuring distribution overlap via the Bhattacharyya coefficient; the resulting dense objective substantially improves grounding accuracy. \textbf{LPO}~\citep{tang2025lpo} (Location Preference Optimization) offers an alternative continuous formulation based on window information entropy and Euclidean distance ($R = R_w \times R_d$), enhancing spatial localization and precision on benchmarks like \textbf{Multimodal Mind2Web}~\citep{deng2023mind2web}.

\paragraph{Combating exploration collapse.}
Even with dense rewards, standard single-sample RLVR can fall into a ``confidence trap'': when the policy is already confident in an incorrect action it never generates the correct one and therefore never receives a positive signal. \textbf{InfiGUI-G1}~\citep{liu2026infiguig1} breaks this deadlock with Adaptive Exploration Policy Optimization (AEPO), which generates multiple candidate answers per forward pass and scores them via an efficiency-derived reward $\eta = U / C$ (accuracy over candidate count). A collinearity penalty further encourages spatial diversity among candidates, creating learning signals for otherwise permanently ``unlearnable'' samples and improving overall semantic alignment.

\subsubsection{LLM-as-Judge Rewards}
\label{sec:llm-judge}

For tasks too ambiguous for closed-form rules, foundation models can serve as generalized reward functions. Recent progress focuses on two intertwined threads: \emph{improving judge accuracy} and \emph{mitigating reward hacking}.

\paragraph{From passive inspection to proactive verification.}
Static LLM judges evaluate fixed trajectory logs passively, often struggling with borderline cases. \textbf{ProRe}~\citep{dai2025prore} addresses this via a reasoner--actor architecture where models decompose evaluation into state-probing tasks executed in the live environment. By gathering active evidence, it significantly improves reward precision and downstream success rates. Similarly, \textbf{SmartSnap}~\citep{cai2025smartsnap} embeds verification directly into the agent's objective: agents are trained to both complete tasks and capture curated visual evidence, allowing lightweight judges to evaluate specific snapshots rather than full, noisy trajectories.

\paragraph{Reducing false positives and reward hacking.}
Enhancing the reliability of judge signals is critical. \textbf{ZeroGUI}~\citep{yang2025zerogui} employs a multi-query unanimous-agreement voting mechanism on trajectory screenshots to drastically reduce false-positive rates and self-hallucinations. To address reward hacking at its root, \textbf{WebRL}~\citep{qi2024webrl} trains its reward model on on-policy trajectories. This couples reward updates to the policy's evolving distribution, mitigating the mismatch exploited by adversarial actions. Nonetheless, maintaining judge robustness under sustained policy optimization remains an open challenge.

\subsubsection{Learned Rewards}
\label{sec:learned-rewards}

Learned reward functions occupy a middle ground: more flexible than hand-crafted rules, more sample-efficient than LLM judges. In the GUI domain they have been most impactful for spatial grounding, where the geometry of the interface provides a natural inductive bias. GUI-G$^{2}$'s Gaussian framework (Section~\ref{sec:rule-based}) doubles as a learned reward once its adaptive variance $\sigma \propto$ element size is treated as a geometry-conditioned function rather than a fixed hyperparameter \citep{tang2025guig2}: small elements (e.g., close-button icons) receive a tight reward landscape while large elements (e.g., banner images) receive a broad one---a distinction critical in high-resolution interfaces where targets can span fewer than $10\times10$ pixels. InfiGUI-G1's Adaptive Exploration Reward \citep{liu2026infiguig1} goes further by making the reward a function of the \emph{full candidate set} rather than a single prediction; the group-relative structure that distinguishes AEPO from na\"{\i}ve best-of-$N$ reranking connects it directly to the GRPO family of algorithms.

Adjacent VLM reward-modeling work also matters even when it originates outside GUI automation. \textbf{MARVL}~\citep{zhou2026marvl}, though robotics-focused, highlights issues that transfer directly to screen agents: learned visual rewards can mis-ground spatial relations, overfit to superficial visual cues, or be exploited by policies that optimize the evaluator rather than the task. Its remedies---stronger spatial grounding, adversarial reward validation, and tighter coupling between perception and action evidence---suggest how GUI reward models can move beyond screenshot-level plausibility toward robust process evaluation.

\subsection{Data Efficiency}
\label{sec:data-eff}

Online RL in live GUI environments is computationally expensive. Page rendering and network operations severely limit the throughput of environment interactions, making standard, data-hungry RL algorithms slow. Consequently, maximizing \emph{data efficiency}---the policy improvement per environment interaction---is a central objective. Three complementary strategies have emerged to address this bottleneck: synthetic data generation via world models (increasing effective data volume at lower cost), enhancement of existing human demonstrations (improving signal quality per sample), and iterative self-improvement loops that recycle the agent's own experience. A complementary systems-level direction is automated training design: \textbf{AutoRL}~\citep{afshar2022automated} suggests that hyperparameters, curricula, and even architecture choices can themselves become optimization targets, which is especially valuable when each GUI rollout is expensive.

\subsubsection{Synthetic Data via World Models}
\label{sec:world-models}

The core idea is to replace or substantially supplement expensive real-environment rollouts with trajectories generated by surrogate models. Two complementary approaches have emerged. At the \emph{reasoning} level, \textbf{DreamGym}~\citep{chen2025dreamgym} distills environment dynamics into an abstract textual state space, using chain-of-thought reasoning over retrieved real trajectories to simulate experiences. Combined with an adaptive task curriculum, it enables previously infeasible online RL on complex web benchmarks. At the \emph{action} level, \textbf{UI-Simulator}~\citep{wang2025uisim} employs the LLM itself as a world simulator to predict visual or textual outcomes directly without rendering interfaces, achieving comparable performance to larger models with significantly less data. These strategies demonstrate that surrogate trajectories can efficiently match the learning value of real rollouts. Complementary methods include \textbf{SimURA}~\citep{deng2025simura}, \textbf{WebSynthesis}~\citep{gao2025websynthesis}, \textbf{WebWorld}~\citep{xiao2026webworld}, and \textbf{Code2World}~\citep{zheng2026code2world}.

\subsubsection{Enhancement of Human Demonstrations}
\label{sec:demo-enhance}

Raw human demonstrations are often noisy and incomplete. Enriching them via structured post-processing or tapping alternative data sources can substantially improve the signal-to-noise ratio.

Structure-based approaches refine existing traces log. \textbf{GUI-ReWalk}~\citep{lin2025guirewalk} converts undirected exploration into targeted RL training data through backward annotation, capturing complex cross-application workflows. Conversely, \textbf{Prune4Web}~\citep{zhang2025prune4web} addresses the DOM-tree size bottleneck by auto-generating Python scripts to prune irrelevant elements, improving grounding accuracy.

An orthogonal direction exploits \emph{novel data sources}. \textbf{Watch-and-Learn}~\citep{song2025watch} trains a dynamics model to predict actions from software usage videos on YouTube, demonstrating that instructional videos can serve as scalable alternatives to step-by-step demonstrations. These results confirm that the \emph{structure} and \emph{diversity} of demonstrations are as critical as their volume. Additional synthesis pipelines include \textbf{AgentTrek}~\citep{xu2024agenttrek} and \textbf{OS-Genesis}~\citep{sun2025genesis}.

\subsubsection{Iterative Self-Improvement}
\label{sec:self-improve}

Rather than relying on fixed datasets, iterative self-improvement allows agents to interact with the environment, collect fresh experience, and update their own training data. Methods vary in their feedback structures. \textbf{Co-EPG}~\citep{zhao2025coepg} features a dual-model (Planner--Grounder) architecture refined via dynamic rewards: the planner learns executable strategies, while the grounder masters low-level intent fulfillment, enabling rapid success on Mind2Web with minimal annotations. Alternatively, \citet{zhang2025agentcpm} iteratively use Implicit World Modeling and Self-Reflection as supervision signals to substantially boost performance across diverse tasks.

\rev{A notable by-product of self-improvement is the appearance of \emph{deliberation-like traces}. As discussed in Section~\ref{sec:offline-representative}, \textbf{GUI-R1}~\citep{luo2025guir1} generates internal planning-style monologues when trained via GRPO on limited samples. The implication is suggestive rather than conclusive: when action spaces and rewards are sufficiently structured, RLVR may reduce reliance on explicit reasoning annotations.}

\subsection{Technical Innovations}
\label{sec:tech-innov}

Beyond reward design and data strategies, a cluster of recent papers introduces algorithmic, perceptual, and memory innovations that address GUI-specific bottlenecks. We organize the discussion by the sub-problem each innovation targets.

\subsubsection{Algorithmic Advances: Exploration and Multi-Turn Optimization}
\label{sec:algo}

Efficient exploration in GUI agents is complicated by the hybrid action space and a sparse reward landscape. Two strategies have emerged: \emph{curriculum-based credit assignment} and \emph{structured exploration architectures}.

On the credit-assignment side, \textbf{WebRL}~\citep{qi2024webrl}'s self-evolving curriculum generates tasks from the agent's own failure set and simplifies them until they fall within its ``Zone of Proximal Development.'' At a finer granularity, the reward-shaping innovations of \textbf{GUI-G$^{2}$}~\citep{tang2025guig2} and \textbf{InfiGUI-G1}~\citep{liu2026infiguig1} demonstrate that the \emph{shape} of the reward landscape accelerates convergence by providing non-zero gradients. \textbf{Agentic Entropy-Balanced Policy Optimization (AEBPO)}~\citep{dong2025agentic} targets rollout-entropy collapse through dynamic entropy-balanced rollout and optimization, significantly improving data efficiency during training. Related RLVR work on \textbf{Introspective LLM}~\citep{zhou2026look} further suggests that exploration can be controlled at the decoding level by learning temperature policies from hidden states rather than relying on fixed sampling heuristics.

On the structural side, \textbf{Nested Browser-Use Learning}~\citep{li2025nested} separates web agent reasoning into an outer loop for tool-integration and an inner loop for in-page goal-driven exploration. This hierarchical decomposition proves remarkably data-efficient without requiring massive sets of synthetic trajectories, highlighting that structuring the exploration process is as important as scaling data.

\subsubsection{Multimodal Perception: Active and Adaptive Visual Grounding}
\label{sec:perception}

A GUI agent that understands \emph{what} to do may still fail if it cannot \emph{see} the correct pixel. Recent innovations attack this perceptual bottleneck through \emph{active perception} and \emph{attention alignment}.

Beyond point localization, \textbf{UILoop}~\citep{li2026s} frames GUI reasoning as a cyclic Screen--UI-elements--Action process, emphasizing explicit learning of element localization, semantic functions, and usage patterns through the 26K-sample UI Comprehension-Bench.

\paragraph{Decoupling System 1 execution and System 2 planning.}
Active perception converts grounding into an iterative process. \textbf{GUI-Eyes}~\citep{chen2026guieyes} lets the agent autonomously invoke visual tools (crop, zoom) before making coordinate predictions, with tool-use decisions learned efficiently through GRPO. In grounding experiments from \textbf{GUI-G$^{2}$}~\citep{tang2025guig2} and \textbf{InfiGUI-G1}~\citep{liu2026infiguig1}, explicit Chain-of-Thought (CoT) reasoning lowers coordinate-prediction accuracy under the tested conditions. This performance gap reveals task-specific tradeoffs between direct grounding and deliberative planning. It shows that direct action heads can outperform text-mediated spatial grounding in some scenarios, while planning and execution do not need strict decoupling. Consequently, a \emph{System 1}/\emph{System 2} split serves as a task-conditional architectural option rather than a universal rule.

Attention-alignment methods attack the same problem from the model-internals side. \textbf{GUI-AIMA}~\citep{zhou2025gui} designs patch-level labels and an \texttt{<ANCHOR>} token for intrinsic multimodal attention alignment. Alternatively, \textbf{GUI-Actor}~\citep{wu2025gui} bypasses explicit coordinate regression entirely by predicting a heatmap directly on the feature map through an attention-driven action head. Together, these approaches demonstrate that visual grounding is substantially improved by reshaping \emph{how} the model attends.

\subsubsection{Memory and Planning: Sustaining Context over Long Horizons}
\label{sec:memory}

\rev{GUI tasks are partially observable from the agent's perspective, so the next action often depends on information distributed across the interaction history. Effective solutions selectively compress the screenshot and metadata history, treating memory management as a \emph{learned behavior} jointly optimized with the policy. Emerging approaches include \textbf{MemR}~\citep{du2025memr} for memory modeling, \textbf{MemSearcher}~\citep{yuan2025memsearcher} for memory retrieval, \textbf{auto-scaling continuous memory}~\citep{wu2025auto}, \textbf{ELMUR}~\citep{cherepanov2025elmur} for extending effective horizons in partially observable settings, and \textbf{AgentProg}~\citep{tian2025agentprog} for program-guided context management.}

Dynamic textual compression is the most direct strategy. \textbf{WebAgent-R1}~\citep{wei2025webagent} has the agent output a textual summary alongside each action to drastically reduce token consumption. \textbf{MGA}~\citep{cheng2025mga} further structures this idea through independent context state triplets managed by an Abstract Memory Agent. Similarly, \textbf{MAGNET}~\citep{sun2026magnet} constructs a memory-driven knowledge evolution framework that dynamically updates a skill library from environmental feedback. \textbf{HAR}~\citep{wang2025history} complements these strategies with a reflective learning process and a Think-More-Than-Step policy that explicitly re-examines past decisions before acting.

The overarching principle is that memory compression is not a pre-processing step but a learned capability. For long-horizon tasks, \textbf{Plan-and-Act}~\citep{erdogan2025plan} explicitly separates planning from execution to sustain coherent behavior. In GUI automation, reward design, data collection, perception, and memory management form a coupled system that determines whether the agent can sustain coherent behavior in complex real-world tasks.

\section{Training Resources}
\label{sec:training-resources}

Robust training of RL-based GUI agents requires a comprehensive ecosystem spanning interactive environments for policy learning, large-scale datasets for pre-training, and specialized frameworks for implementing RL algorithms on multimodal models. This section provides a systematic overview of the training resource landscape that underpins the RL paradigms (Section~\ref{sec:rl-methods}) and cross-cutting dimensions (Section~\ref{sec:key-dimensions}) discussed above. Rather than re-describing algorithmic innovations, we focus on the characteristics of each resource and its role in the RL training pipeline.

\begin{figure}[h]
    \centering
    \includegraphics[width=0.93\textwidth]{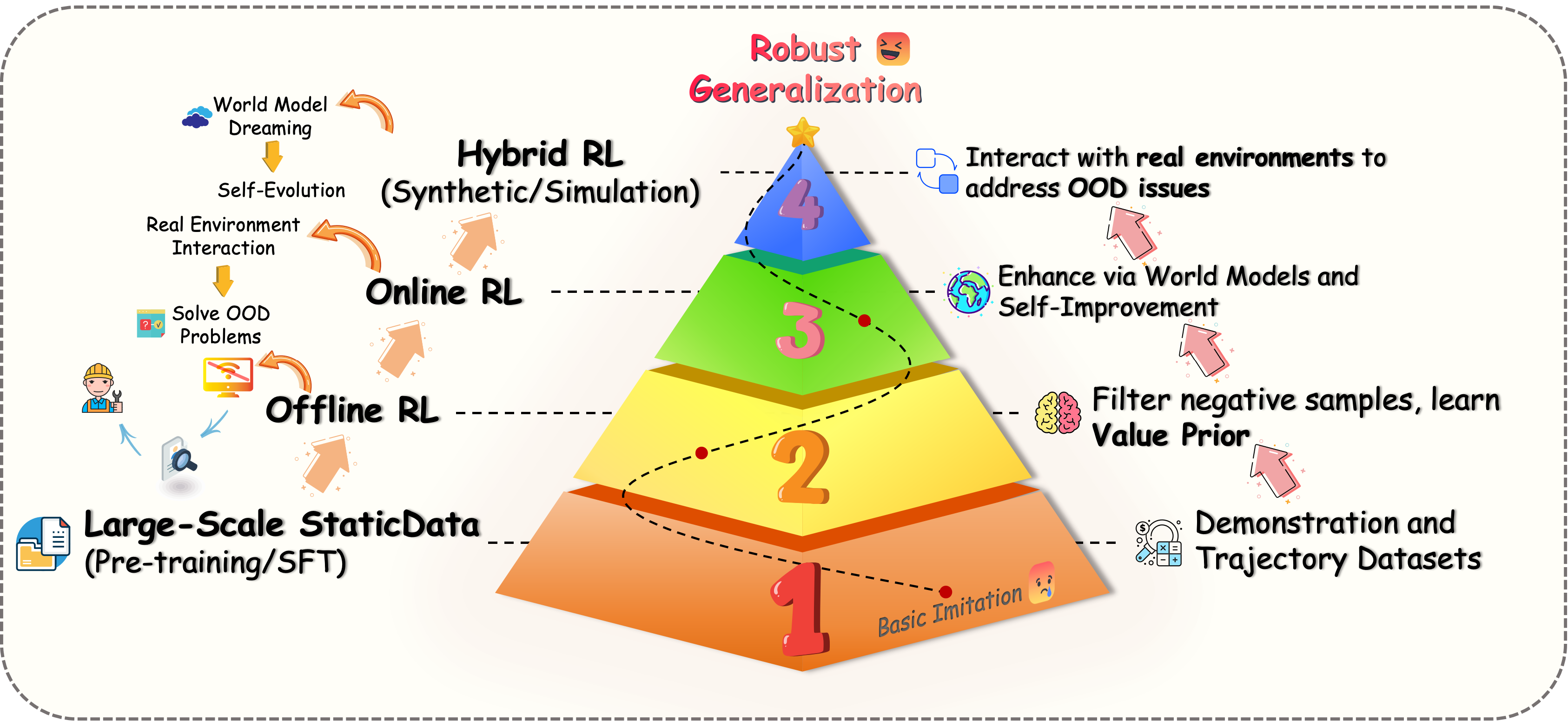}
    \caption{This pyramid depicts a four-stage data-training pipeline for agent capability, progressing from static data imitation to offline RL, synthetic simulation, and online RL, to achieve robust generalization.}
    \label{fig:training-funnel} 
\end{figure}

\subsection{Datasets}
\label{sec:datasets}

The efficacy of RL-based GUI agents fundamentally depends on the quality and diversity of training data. Unlike traditional supervised learning paradigms that prioritize scale and human similarity, RL-centric datasets must satisfy distinct requirements: completeness of state representations (for Critic networks), density of reward signals (for sparse reward mitigation), and environmental interactivity (for large-scale online exploration). This section systematically categorizes the data landscape into three strategic dimensions that collectively enable the RL training pipeline—from policy cold-start to self-evolution.

\subsubsection{Demonstration and Trajectory Datasets}

Demonstration datasets serve as the cornerstone for policy initialization and offline RL, addressing the fundamental challenge of exploration in high-dimensional action spaces where pixel-level click actions can reach millions of possibilities. These datasets vary significantly in their structural characteristics, each offering distinct advantages for different RL training requirements. Table~\ref{tab:demo-datasets} summarizes the key demonstration and trajectory datasets.

\begin{table}[t]
\caption{Demonstration and trajectory datasets for policy initialization and offline RL.}
\label{tab:demo-datasets}
\centering
\small
\begin{tabular}{llll}
\toprule
\textbf{Dataset} & \textbf{Platform} & \textbf{Scale} & \textbf{RL Role} \\
\midrule
AitW & Mobile & 715K traj. & Offline pre-training \\
AndroidControl & Mobile & --- & Hierarchical credit assignment \\
Mind2Web & Web & 2K+ tasks & Action space compression \\
OmniACT & Desktop/Web & 9.8K tasks & Macro-action generation \\
\bottomrule
\end{tabular}
\end{table}

For mobile platforms, \textbf{Android-in-the-Wild (AitW)}~\citep{rawles2023androidinthewild} provides the largest publicly available corpus with 715K trajectories spanning the Google application ecosystem. Its visual-only dependency—lacking DOM or accessibility trees—forces agents to develop robust pure-vision policies, proving advantageous for generalization to real-world applications (games, Flutter/Unity apps) that deny structured UI access. Notably, AitW's inclusion of human operation noise (mis-taps, hesitations, failed swipes) transforms traditional liabilities into assets for offline RL: through \textit{Advantage Weighting} techniques, models learn to distinguish high-value from low-value actions by mining sub-optimal trajectories, as demonstrated by DigiRL's successful offline pre-training (Section~\ref{sec:hybrid-representative}). In contrast, \textbf{AndroidControl}~\citep{li2024effects} prioritizes structural richness over scale, introducing hierarchical instruction annotations that directly address the credit assignment problem. By providing both high-level intents and intermediate low-level instructions (e.g., ``open clock app,'' ``tap add button''), it enables \textit{Hierarchical Reinforcement Learning (HRL)} where sparse terminal rewards decompose into dense step-wise signals. Its complete XML View Hierarchy metadata further supports structured state representations via Graph Neural Networks—critical for stable value function estimation in offline algorithms like IQL and CQL.

Web environments present fundamentally different challenges, as action spaces are inherently discrete (selecting DOM elements) rather than continuous. \textbf{Mind2Web}~\citep{deng2023mind2web} addresses this by spanning 2000+ tasks across 137 websites with comprehensive DOM tree annotations, enabling training of efficient \textit{Grounding Models} that compress action spaces from $O(10^4)$ to $O(10^1)$ candidates via semantic filtering—making RL optimization computationally feasible. Its cross-site diversity enforces learning of HTML tag semantics (e.g., \texttt{<input type="search">} universally indicates search functionality) rather than brittle coordinate memorization, yielding policies that generalize beyond training domains. For desktop environments, \textbf{OmniACT}~\citep{kapoor2024omniact} pioneered the \textit{Action-as-Code} paradigm where agents generate executable Python scripts (PyAutoGUI) rather than atomic actions. This structured action space fundamentally alters the RL time horizon: instead of executing dozens of fragile atomic clicks, agents output coherent macro-action scripts, shortening episode lengths and enabling more effective sparse reward propagation across its 9802 desktop/web tasks.

\subsubsection{Perception and Grounding Datasets}

Accurate perception forms the sensory foundation for RL value networks and reward models—in open GUI environments lacking API-level feedback, visual understanding is the sole mechanism for state evaluation and task completion verification. These grounding datasets train the ``judges'' that enable RL agents to assess their own performance, with applications spanning reward shaping, hallucination reduction, and state compression. Table~\ref{tab:grounding-datasets} provides an overview of the principal perception and grounding datasets.

\begin{table}[t]
\caption{Perception and grounding datasets for reward shaping and state evaluation.}
\label{tab:grounding-datasets}
\centering
\small
\begin{tabular}{llll}
\toprule
\textbf{Dataset} & \textbf{Platform} & \textbf{Scale} & \textbf{RL Role} \\
\midrule
ScreenSpot Series & Cross-platform & --- & IoU-based reward shaping \\
Ferret-UI & Mobile & --- & Visual CoT; anti-hallucination \\
Rico & Mobile & 66K screens & Grounding pre-training \\
Screen2Words & Mobile & 112K pairs & State compression \\
Widget Captioning & Mobile & 163K caps. & Semantic state abstraction \\
UIGuard & Cross-platform & --- & Safety reward signals \\
\bottomrule
\end{tabular}
\end{table}

The \textbf{ScreenSpot Series (V1/V2/Pro)}~\citep{li2025screenspot} has emerged as the gold standard for visual grounding precision, directly enabling \textit{reward shaping} in RL pipelines. In GRPO training loops (e.g., UI-R1), when an agent outputs a click action $(x, y)$ without immediate environmental feedback, a reward model fine-tuned on ScreenSpot computes the Intersection-over-Union (IoU) between predicted coordinates and ground-truth UI elements, yielding dense signals that guide policy optimization without requiring environmental interaction during early training phases. ScreenSpot-Pro's high-resolution challenges specifically target modern MLLM hallucination issues at production scales. Complementing coordinate-level precision, \textbf{Ferret-UI}~\citep{you2024ferret,li2024ferret} tackles mobile-specific visual reasoning through \textit{any-resolution} adaptability—standard vision encoders (CLIP's $224 \times 224$ squares) severely distort mobile screenshots' elongated aspect ratios. Ferret-UI's fine-grained regional annotations enable \textit{Visual Chain-of-Thought (CoT)} capabilities where agents verbalize spatial reasoning before acting, measurably reducing hallucination behaviors during RL exploration. Similarly, \textbf{Rico}~\citep{deka2017rico} provides rich UI element annotations for Android interfaces that support both grounding model pre-training and UI understanding tasks.

Beyond localization, semantic understanding datasets address the critical challenge of \textit{state compression} for long-horizon tasks. \textbf{Screen2Words}~\citep{wang2021screen2words} and \textbf{Widget Captioning}~\citep{li2020widget} convert pixel states into textual descriptions—encoders trained on these datasets compress high-dimensional visual states into concise semantic summaries (e.g., ``login page with username/password fields''), enabling RL agents to maintain textual memory of multi-page workflows while reserving pixel-level processing for the current frame only. This hybrid multimodal state representation is essential for scaling RL to tasks spanning dozens of interface transitions. For safety-critical applications, \textbf{UIGuard}~\citep{chen2023unveiling} provides dark pattern detection annotations—UI designs that mislead users toward unintended actions. These negative samples enable construction of \textit{safety reward functions} that impose penalties when agents attempt interaction with deceptive elements, implementing Safe RL principles for production deployment.

\subsubsection{Synthetic and RL-Generated Corpora}

The frontier of RL-based GUI agents has shifted toward \textit{``environment-as-data''} paradigms where agents generate unbounded training curricula through autonomous interaction—static datasets, regardless of scale, suffer from distribution mismatch that self-generated on-policy data eliminates. This category encompasses both interactive environments that enable online reinforcement learning and frameworks that synthesize reasoning-augmented trajectories.

\begin{table}[t]
\caption{Synthetic and RL-generated corpora for on-policy data generation and self-improvement.}
\label{tab:synthetic-datasets}
\centering
\small
\begin{tabular}{llll}
\toprule
\textbf{Dataset} & \textbf{Platform} & \textbf{Scale} & \textbf{RL Role} \\
\midrule
AndroidWorld & Mobile & 116 tasks & Ground-truth reward signals \\
WebArena & Web & 812 tasks & Executable state verification \\
VisualWebArena & Web & 910 tasks & Visual reward verification \\
OSWorld & Desktop & 369 tasks & Cross-app state tracking \\
GUI-Bee & Cross-platform & --- & Entropy-driven exploration \\
Explorer & Web & 94K+ traj. & Instruction reverse-engineering \\
UI-TARS & Cross-platform & --- & Reasoning-augmented triplets \\
GUI-R1 & Cross-platform & 3K samples & Emergent reasoning patterns \\
\bottomrule
\end{tabular}
\end{table}

Dynamic interactive environments form the foundation of this paradigm. \textbf{AndroidWorld}~\citep{rawles2024androidworld} functions as the definitive ``Gymnasium'' for mobile RL, generating millions of task variants through parameterized templates (e.g., ``add contact'' with randomized names/numbers) that prevent rote memorization. Its \textit{non-invasive state inspection} via ADB interfaces delivers 100\% accurate ground-truth reward signals by querying Android's underlying SQLite databases—enabling agents like \textbf{AppAgent}~\citep{zhang2025appagent} and \textbf{UI-TARS}~\citep{qin2025ui} to perform millions of trial-and-error interactions that form self-improving data flywheels. For web environments, \textbf{WebArena}~\citep{zhou2023webarena} and \textbf{VisualWebArena}~\citep{koh2024visualwebarena} provide self-hostable simulated internet platforms with executable verification—running backend scripts to validate database state changes rather than shallow HTML comparison. Additional web benchmarks include \textbf{WebCanvas}~\citep{pan2024webcanvas} for online evaluation, \textbf{BearCubs}~\citep{song2025bearcubs} for web agent benchmarking, \textbf{WebWalker}~\citep{wu2025webwalker} for LLM-based web traversal, \textbf{Agent-X}~\citep{ashraf2025agent} for evaluating deep reasoning, and \textbf{TheAgentCompany}~\citep{xu2024theagentcompany} for real-world enterprise tasks. Beyond browsing-based approaches, \textbf{\citet{song2025beyond}} explored API-based web agents, while end-to-end navigation with \textbf{VLMs}~\citep{goetting2024end} demonstrated direct visual navigation capabilities. \textbf{WebAgent-R1} demonstrated that synthetic success trajectories generated through parallel exploration in WebArena outperform human demonstrations, as self-generated data reflects agents' actual capability boundaries. \textbf{OSWorld}~\citep{xie2024osworld} extends this paradigm to desktop environments, providing file system state tracking and cross-application workflow support essential for complex multi-app tasks.

Complementing task-directed environments, exploration-focused approaches address cold-start and generalization challenges. \textbf{GUI-Bee}~\citep{fan2025gui} introduces \textit{exploration data}—trajectories generated via entropy-maximizing autonomous exploration rather than goal completion. Through Q-ICRL mechanisms, agents construct exploration graphs mapping state transition structures and navigation dead-ends, enabling zero-shot adaptation to unseen applications (analogous to humans casually familiarizing themselves with new software). The \textbf{Explorer}~\citep{pahuja2025explorer} framework employs multi-agent pipelines where ``explorers'' randomly walk through web environments discovering novel states while ``annotators'' reverse-engineer natural language instructions, synthesizing 94K+ high-quality trajectories covering long-tail scenarios.

Most recently, reasoning-augmented data generation has emerged as a critical frontier inspired by DeepSeek-R1's success. \textbf{UI-TARS}~\citep{qin2025ui} and \textbf{AutoPlay}~\citep{ramrakhya2025autoplay} generate \textit{(State, Thought, Action)} triplets where teacher models (e.g., GPT-4o) produce detailed reasoning chains (``I need to click the search bar because the desired product isn't visible on the homepage...''), with lightweight verifiers filtering logically inconsistent samples. These datasets enable training of \textit{Process Reward Models (PRMs)} that reward intermediate reasoning validity—not just final outcomes—guiding agents from blind trial-and-error toward logical problem decomposition. The resulting data flywheel---iteratively generating, filtering, and fine-tuning on reasoning traces---has propelled continuous SOTA improvements, as exemplified by \textbf{GUI-R1}'s extreme data efficiency with merely 3K curated samples (Section~\ref{sec:offline-representative}).

\subsection{Interactive Environments}
\label{sec:environments}

Unlike static benchmarks used primarily for evaluation, training environments for online RL must support the standard Markov Decision Process (MDP) interface and efficient resetting mechanisms. The evolution of GUI RL environments has progressed from synthetic sandboxes toward high-fidelity digital twins of real-world interfaces, addressing specific challenges in state representation, action space design, and reward engineering.

\subsubsection{Web and Browser Environments}

Web environments benefit from standardized rendering protocols (HTML/CSS/JS), evolving from synthetic micro-tasks to full internet simulations. Early foundational work like \textbf{MiniWoB++}~\citep{liu2018miniwob} isolated interaction primitives in HTML5 sandboxes, exposing dual modalities but imposing extremely sparse rewards that demanded distributed training solutions like \textbf{CC-Net}~\citep{humphreys2022ccnet}.

The field subsequently shifted toward complex semantic understanding. \textbf{WebShop}~\citep{yao2022webshop} simulated a vast e-commerce platform with dense attribute-overlap rewards, successfully demonstrating sim-to-real transfer. Modern approaches operate on real-world snapshots: \textbf{Mind2Web}~\citep{deng2023mind2web} preserves webpage states for deterministic replay and compresses DOM action spaces via semantic filtering. \textbf{WebArena}~\citep{zhou2023webarena} advances this with self-hostable platforms supporting executable verification, powering frameworks like \textbf{WebRL} (Section~\ref{sec:online-representative}) to utilize self-evolving curricula and Outcome-supervised Reward Models. To unify this fragmented landscape, \textbf{BrowserGym}~\citep{chezelles2024browsergym} aggregates major benchmarks (\textbf{WebArena}, \textbf{MiniWoB++}, \textbf{VisualWebArena}~\citep{koh2024visualwebarena}, \textbf{WebChoreArena}~\citep{miyai2025webchorearena}) under a standardized API that encapsulates critical infrastructure.

\subsubsection{Desktop and OS Environments}

Desktop environments introduce higher-dimensional challenges like file management and multi-app switching. Their engineering foundation relies on headless virtualization: \textbf{OSWorld}~\citep{xie2024osworld} utilizes Docker and QEMU with virtual framebuffers to parallelize environments for massive sample collection. Crucially, it employs execution-based evaluation to inspect side-effects directly (e.g., file state changes), neutralizing hallucination issues common in text-matching and revealing a substantial human-machine gap.

Optimizing these environments focuses on action space design and training stability. \textbf{ComputerRL}~\citep{lai2025computerrl} introduces an \textit{API-GUI hybrid action paradigm} to leverage both API determinism and GUI universality, combating entropy collapse during long-horizon tasks via an interleaved RL and SFT \textit{Entropulse} strategy. For perception robustness, \textbf{ScreenAgent}~\citep{niu2024screenagent} uses VNC protocols for pure pixel-stream control independent of Accessibility APIs, adopting a self-correcting \textit{Plan-Act-Reflect} loop. Ultimately, scaling desktop environments depends on infrastructure maturity alongside algorithmic innovation.

\subsubsection{Mobile Environments}

Mobile platforms require specialized environments due to dense, gesture-based controls and isolated app ecosystems. \textbf{AndroidEnv}~\citep{toyama2021androidenv} provides a standard MDP, modeling continuous virtual finger movements that extend time horizons and complicate credit assignment. Due to standard emulators' high resource costs, recent tools like \textbf{UISim}~\citep{xiang2025uisim} offer streamlined image-based UI simulators, typically deployed with massive parallelization.

Task and reward designs have also advanced. \textbf{AndroidWorld}~\citep{rawles2024androidworld} prevents overfitting through dynamic task parameterization and extracts zero-noise rewards by directly querying system internals. To combat reward sparsity, \textbf{Mobile-Env}~\citep{zhang2023mobile} employs background evaluators to monitor system states and provide dense intermediate feedback.

To overcome covariate drift inherent in static behavioral cloning, the field is shifting toward dynamic interaction paradigms. \textbf{DigiRL} (Section~\ref{sec:hybrid-representative}) combines offline and online RL. Modern infrastructures easily scale this paradigm by decoupling CPU simulation from GPU inference (e.g., \textbf{MAI-UI}~\citep{zhou2025mai}, \textbf{MobileGUI-RL}~\citep{shi2025mobilegui}), often leveraging GRPO with efficiency rewards to train agents that discover concise operational paths.

\subsubsection{Cross-Platform Trends and Synthesis}

Several cross-cutting trends emerge from this environmental evolution. First, \textit{visual-first state representation} is becoming dominant: as structured metadata access is increasingly precluded, convergence is driven toward pixel-first approaches (e.g., \textbf{ScreenAgent}) that prioritize cross-platform generality. Second, \textit{action spaces are evolving} from raw coordinates toward API-GUI hybrids (\textbf{ComputerRL}) or coordinate-free semantic grounding (\textbf{GUI-Actor}). Recent benchmarks like \textbf{OSWorld-MCP}~\citep{jia2025osworld} and \textbf{MCPWorld}~\citep{yan2025mcpworld} formalize tool invocation and hybrid API/GUI evaluation. Third, \textit{reward engineering} has progressed from binary system state verification (\textbf{AndroidWorld}, \textbf{OSWorld}) to learned Outcome-supervised Reward Models (\textbf{WebRL}); as task complexity increases, deterministic verification becomes infeasible, driving the adoption of RLAIF paradigms. Finally, \textit{Sim-to-Real gaps} persist; successful transfer pipelines combine offline pretraining, simulated fine-tuning, and real-world deployment (e.g., \textbf{DigiRL}), with domain randomization emerging as a critical technique.

\subsection{RL Infrastructure and Tools}
\label{sec:infrastructure}

Constructing scalable training loops for Multimodal LLM-based GUI agents requires specialized infrastructure addressing critical computational bottlenecks unique to this domain. Unlike text-only RL systems where training is typically compute-bound, GUI agent training transitions to \textit{I/O-bound} workloads: environment rendering, screenshot transmission, and perception processing dominate execution time, making traditional synchronous RL frameworks inefficient. The emerging infrastructure ecosystem comprises four interdependent dimensions: VLM-RL algorithm libraries optimizing for multimodal inputs, distributed architectures decoupling rollout and training, reward engineering tools solving the sparse/deceptive feedback problem, and memory management systems enabling long-horizon coherent reasoning.

\subsubsection{VLM-RL Algorithm Libraries and Framework Evolution}

The foundational algorithmic layer has evolved from generic RLHF libraries \citep{hu2024openrlhf} toward GUI-specialized implementations balancing computational efficiency with learning signal quality.

\textbf{GRPO and Memory-Efficient Policy Optimization:} Group Relative Policy Optimization (GRPO) has emerged as the dominant algorithm for GUI RL, employed in systems like \textbf{Mano}, \textbf{GUI-R1}, \textbf{InfiGUI-G1}, and \textbf{GUI-Eyes}. GRPO eliminates the memory-intensive Critic network, a critical optimization for long visual sequences; it achieves advantage estimation across trajectory groups instead of learned value functions:
\[
\hat{A}_i = \frac{r_i - \text{mean}(\mathbf{r})}{\text{std}(\mathbf{r})}, \qquad \mathcal{L}_{\text{GRPO}}(\theta) = -\frac{1}{G}\sum_{i=1}^{G} \min\!\left(\rho_i \hat{A}_i,\; \text{clip}(\rho_i, 1\!-\!\epsilon, 1\!+\!\epsilon)\hat{A}_i\right) + \beta\, D_{\text{KL}}(\pi_\theta \| \pi_{\text{ref}})
\]
where $\rho_i = \pi_\theta(o_i|x)/\pi_{\text{old}}(o_i|x)$. Libraries like \textbf{veRL}~\citep{sheng2025hybridflow} and \textbf{ReaL}~\citep{mei2025real} provide production-grade GRPO implementations. They achieve dramatic throughput improvements through \textit{HybridFlow}~\citep{sheng2025hybridflow}, a decoupled paradigm enabling Actor models to transition seamlessly between inference backends (\textbf{vLLM}, built on PagedAttention~\citep{kwon2023efficient}) and training frameworks (\textbf{Megatron-LM}~\citep{shoeybi2019megatron}, \textbf{FSDP}~\citep{zhao2023pytorch}) without redundant weight replication.

\textbf{Hybrid Offline-to-Online Frameworks:} To address sample inefficiency during cold-start phases, frameworks increasingly support offline-to-online transitions. \textbf{OpenRLHF}~\citep{hu2024openrlhf} implements Advantage-Weighted Regression (AWR) to extract behavioral priors from static demonstrations before online refinement. \textbf{DigiRL} (Section~\ref{sec:hybrid-representative}) exemplifies this by combining offline initialization with online fine-tuning via instruction-level value functions.

\textbf{Generative and Reasoning-Augmented Reward Models:} Emerging frameworks integrate generative reward modeling, where VLMs directly compare states. \textbf{RewardDance}~\citep{wu2025rewarddance} reformulates this as a binary classification task (``Is state $s_A$ better than state $s_B$?''), using the positive token's log-probability as the reward:
\[
r = \log P(\text{``Yes''} \mid s_A, s_B, \mathcal{O})
\]
This approach utilizes foundation models' full representational capacity and exhibits robust resistance to reward hacking.

\subsubsection{Distributed Rollout and Training Architectures}

\begin{figure}[h]
    \centering
    \includegraphics[width=0.93\textwidth]{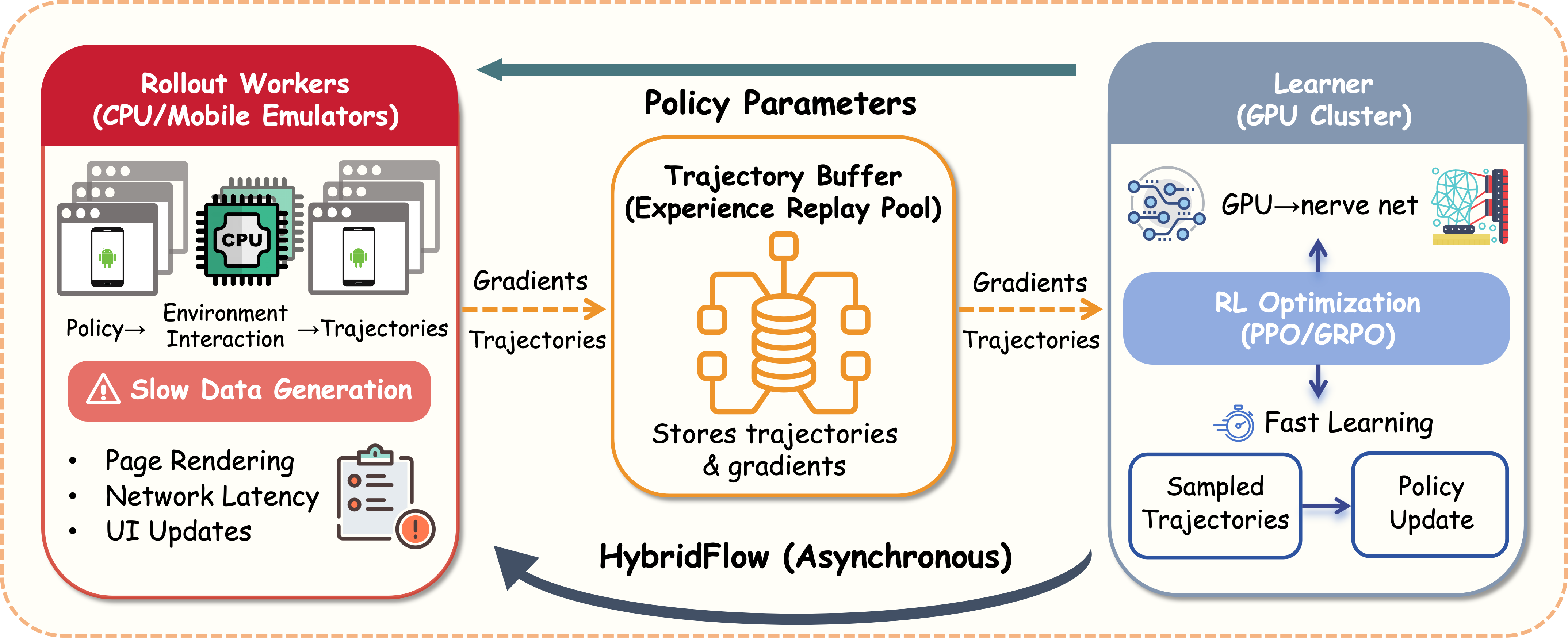}
    \caption{An asynchronous distributed architecture for GUI RL agent training, addressing slow environment interaction latency. It decouples slow data generation (CPU/mobile emulators as Rollout Workers) from fast GPU-cluster learning. Rollout Workers feed trajectories/gradients to a buffer, while the Learner sends updated policy parameters back asynchronously via HybridFlow, enabling massive parallelism and efficient GPU utilization.}
    \label{fig:training-architecture}
\end{figure}

The infrastructure transition from synchronized training to fully asynchronous architectures represents a fundamental paradigm shift driven by I/O latency. A single GUI environment step (screenshot capture, OCR, action execution, rendering) commonly takes 0.5--2 seconds, rendering synchronous PPO inefficient: GPUs remain idle during rollout phases while rollout workers stall during training phases. Modern systems decouple these workloads entirely.

\textbf{Fully Asynchronous Actor-Trainer Separation:} AReaL~\citep{fu2025areal} pioneered production-scale asynchronous RL for long-horizon agent tasks. Its architecture partitions workflows into decoupled \textit{Rollout Workers} (environment-facing processes continuously sampling trajectories) and \textit{Trainer Workers} (GPU-centric processes consuming data from replay buffers). This decoupling introduces \textit{data staleness}—when trainers update policy $\pi_{\theta_{t+1}}$, rollout workers may still use $\pi_{\theta_t}$—yet AReaL's algorithm-system co-design mitigates this through PPO variants tolerant of stale data. Benchmark results demonstrate 3× acceleration over synchronous systems, with linear scaling to 1000+ GPUs. The architectural pattern is now standard across research and production systems.

\textbf{Heterogeneous Hardware Scheduling:} HETHUB~\citep{xu2024hethub} addresses the practical reality that GPU clusters are rarely homogeneous. Modern datacenters mix NVIDIA A100 (high-bandwidth, suited for vision encoding), H100 (compute-intensive decoders), and consumer GPUs. HETHUB's automatic parallel planner analyzes hardware specifications and dynamically assigns model components: vision encoders to bandwidth-optimized A100s, transformer layers to compute-dense H100s, etc. This fine-grained scheduling reduces cluster completion time by 30--50\% compared to naive partitioning. DistRL~\citep{wang2024distrl} and Agent.xpu~\citep{wei2025agent} further extend this paradigm to edge and mobile scenarios: DistRL implements centralized training with decentralized rollout across mobile devices, while Agent.xpu schedules antagonistic tasks (low-latency user interaction vs. high-throughput RL training) on System-on-Chip devices through kernel-level preemption.

\textbf{API-GUI Hybrid Action Paradigms:} ComputerRL introduces a critical architectural innovation: permitting agents to choose between low-level GUI actions (pixel coordinates) and high-level system APIs. This hybrid paradigm allows agents to call $\texttt{get\_file\_content(path)}$ instead of laboriously opening a file browser, reducing trajectory length and enabling learning in extremely constrained sample budgets. The unified action interface abstracts platform-specific implementation (Win32 APIs, X11 calls, macOS Cocoa) behind a standard interface, simplifying distributed training across heterogeneous operating systems.

\subsubsection{Reward Engineering and Verification Systems}

Reward design is a critical bottleneck, balancing between uninformative sparse terminal rewards and exploitable hand-crafted dense signals. Modern systems address this tension through multi-layered architectures. To establish reliable feedback, VAGEN~\citep{cui2026agentic} introduces \textbf{Agentic Verification}, replacing passive LLM observers with active environmental probing---such as executing system commands to verify side effects---to enable noise-free Reinforcement Learning from Verifiable Rewards (RLVR). To overcome long-horizon sparsity, \textbf{Process Reward Models} have emerged; ProgRM~\citep{zhang2025progrm}, for instance, automatically extracts intermediate milestones from demonstrations to provide dense progress estimations. Concurrently, methods including InfiGUI-G1~\citep{liu2026infiguig1} and Mano~\citep{fu2025mano} utilize \textbf{Composite Rewards} that integrate multiple constraints---such as IoU, realistic bounding box sizes, and format validity---to prevent reward hacking and mode collapse. Finally, \textbf{Autonomous Evaluation} pipelines, as demonstrated by systems like ZeroGUI, leverage aggregated VLM-as-judge scoring to construct self-evolving, zero-human-cost training curricula.

\subsubsection{Memory Management and Long-Horizon Reasoning}

Extended interaction horizons pose acute challenges to context management: full trajectory concatenation becomes prohibitively expensive, yet lossy compression risks critical information loss. Contemporary systems implement learnable and adaptive memory mechanisms.

\textbf{Reinforcement Learning-Driven Memory Operations:} \textbf{Memory-R1}~\citep{yan2025memory} treats memory management as a learnable decision process. A Memory Manager Agent operates an action space $\{\texttt{ADD}, \texttt{UPDATE}, \texttt{DELETE}, \texttt{NOOP}\}$ on an external memory store. Critically, this Agent only receives reward when updated memory helps solve the downstream task. This outcome-driven training forces agents to actively suppress stale information and consolidate persistent facts, substantially outperforming fixed-length context windows.

\textbf{Hierarchical Working Memory and Chunking:} \textbf{Hi-Agent}~\citep{wu2025hi} (Section~\ref{sec:hybrid-representative}) enforces hierarchical decomposition. Agents first propose subgoals, then execute low-level primitive actions. Once achieved, the system automatically collapses the action sequence into a high-level summary, ensuring context windows retain fine-grained details for current subtasks alongside coarse history summaries. This chunking mechanism extends effective planning horizons while maintaining manageable token counts.

\textbf{Structured State Representation and Memory Triads:} \textbf{Memory-Driven GUI Agent (MGA)}~\citep{cheng2025mga} decouples reasoning from historical artifacts through a three-component state representation, where Spatial Cues represent extracted UI layout information and Structured Memory contains dynamic summaries. This explicitly separates current decisions from historical influences, reducing trajectory collapse caused by historical context pollution.

\subsubsection{Integration and Ecosystem Standardization}

Modern infrastructure increasingly emphasizes interoperability. \textbf{BrowserGym}~\citep{chezelles2024browsergym} provides a unified Gymnasium API aggregating diverse benchmarks (\textbf{WebArena}, \textbf{MiniWoB++}, \textbf{VisualWebArena}) while handling infrastructure concerns like Docker sandboxing and DOM parsing. Similarly, \textbf{GUI-MCP} standardizes tool-calling interfaces, allowing reward functions to universally verify task completion across heterogeneous applications. Open-source scaffolding platforms like \textbf{OpenHands}~\citep{wang2024openhands} and \textbf{AutoGen}~\citep{wu2024autogen} wrap raw LLMs with memory management, tool interfaces, and trajectory logging, reducing engineering friction for practitioners building RL systems.

\section{Challenges and Future Directions}
\label{sec:future}

The preceding sections show that RL already improves GUI agents along three tightly coupled axes: reward design, data efficiency, and long-horizon decision making. Looking forward, the central question is no longer whether RL is useful for GUI automation, but what kind of agents these training paradigms are ultimately producing. In our view, the next stage of the field is not simply ``better tool use,'' but the emergence of agents that can persist, adapt, and act reliably within evolving software ecosystems.

\subsection{Digital Worlds}

We first discuss the conceptual transition from task-specific GUI agents to persistent computer-use agents embedded in broader digital environments.

\subsubsection{Digital Inhabitants}

We use the term \textbf{digital inhabitants} to describe a stronger class of computer-use agents: systems that do not merely execute isolated instructions on a screen, but maintain persistent competence within digital environments. A digital inhabitant should be able to internalize interface regularities, adapt to software updates, accumulate reusable experience across tasks, and operate under stable behavioral constraints over long horizons. This perspective extends GUI agents from one-shot task solvers to continual actors embedded in a broader digital world.

For RL-based GUI agents, this shift is especially natural. As discussed in Sections~\ref{sec:rl-methods} and~\ref{sec:key-dimensions}, GUI environments expose sequential structure, delayed rewards, and partial observability in a form that is difficult to handle with static imitation alone. RL provides the mechanism for converting interaction into competence. More importantly, it offers a path toward agents that learn not only \emph{which} action sequence succeeds once, but \emph{why} certain interface patterns, failure modes, and recovery strategies recur across applications and platforms. In this sense, GUI agents may become the most practical substrate for studying general computer-use agents: they sit at the boundary between narrow application automation and open-ended digital interaction.

\subsubsection{Agent-Native Environments}

At the same time, a fully developed digital inhabitant may not ultimately operate within computers designed primarily for humans. Current GUI agents are trained to act through human-oriented abstractions such as windows, icons, forms, and cursor-level manipulation. This setting is important because it covers the existing software world, but it may also be transitional. In the longer run, the natural endpoint is likely to be \emph{agent-native operating environments}: operating systems, execution substrates, and even hardware interfaces designed explicitly for machine actors rather than retrofitted from human-computer interaction.

From this perspective, today's GUI agents play a dual role. In the short term, they are practical automation systems for the legacy digital ecosystem. In the long term, they are a bridge technology that reveals which components of computer use should remain embodied and interactive, and which should be re-designed into machine-readable primitives. The future infrastructure of agent society may therefore include persistent agent identities, explicit permission and accountability layers, auditable action logs, machine-native task protocols, and regulatory rules that define what an autonomous agent is allowed to perceive, remember, exchange, and execute. Only within such infrastructure can computer-use agents become true digital inhabitants rather than highly capable users of human software.

To make this bridge actionable for practitioners and standards bodies, we see three near-term standardization targets. First, interfaces should expose \emph{machine-readable UI schemas} that provide stable semantics for roles, affordances, constraints, and state transitions beyond raw pixels. Second, platforms should define \emph{verifiable outcome APIs} that expose task completion evidence, side-effect traces, and policy-compliance checks in a form that can be used directly by reward and evaluation pipelines. Third, the community needs \emph{reference sandbox specifications} that formalize permission scopes, rollback behavior, logging requirements, and human override mechanisms, so that training and deployment can be compared under shared safety assumptions.

\subsection{Technical Roadmap}

The next set of challenges concerns the technical conditions under which RL can support robust, scalable, and generalizable computer-use behavior.

\subsubsection{Reward Interfaces}

One of the clearest lessons of this survey is that verifiability is both the main opportunity and the main bottleneck for RL in GUI agents. Rule-based rewards, LLM-as-judge signals, and learned reward models (Section~\ref{sec:key-dimensions}) have made it possible to train agents on increasingly realistic tasks, yet each remains incomplete. The difficulty is that real computer-use goals are rarely exhausted by terminal success predicates. Tasks such as purchasing, scheduling, document editing, or enterprise workflow execution require semantic correctness, procedural compliance, and often user-specific preferences, none of which are fully captured by URL changes or form submission events.

This suggests an important future direction: reward design must move from \emph{task completion} toward \emph{intent satisfaction under constraints}. For GUI agents, that means richer evaluation pipelines that jointly assess outcome quality, process correctness, and recoverability after mistakes. For computer-use agents more broadly, it implies that RLVR will increasingly depend on layered evaluators, combining executable checks, environment feedback, and model-based judgment. The open problem is not simply building stronger reward models, but constructing reward interfaces that remain reliable as the task space expands from benchmark-style episodes to open-world digital work.

\subsubsection{I/O-Constrained Learning}

As argued in Section~\ref{sec:data-eff}, environment interaction in GUI settings is fundamentally slow. Rendering, network delay, and screenshot transmission make online RL expensive in a way that is qualitatively different from classic simulators. This I/O wall is not just an engineering nuisance; it is a structural constraint that shapes which learning strategies are viable. It explains why progress has increasingly relied on hybrid pipelines that combine demonstrations, offline optimization, selective online exploration, and synthetic data generation.

Table \ref{tab:gui_throughput} makes this constraint more operational. The numbers are best read as order-of-magnitude planning ranges than fixed benchmark results, since latency depends on browser engine, emulator density, network locality, screenshot resolution, and reward instrumentation. Even under optimistic assumptions, a single live GUI environment usually produces only sub-Hz to low-Hz interaction streams, while parallel rollout primarily hides I/O stalls rather than eliminating them.

The practical implication is that scaling online GUI RL is rarely a matter of adding GPUs alone. Training systems must either increase environment multiplicity through asynchronous rollout workers, reduce per-step observability costs through state compression and pruning, or shift more exploration into surrogate dynamics before periodically re-grounding on live interfaces.

This also makes automation of the training stack increasingly important. AutoRL-style methods \citep{afshar2022automated} can reduce manual search over curricula, reward weights, rollout schedules, and model sizes, while infrastructure-centered systems such as AgentCPM-Explore \citep{chen2026agentcpm} show that reward denoising and context compression are first-order design choices under real browser and app I/O noise.

\begin{table}[t]
\centering
\caption{Representative GUI step-time bottlenecks and rollout throughput across common training settings (environment-side latency only; model inference excluded).}
\label{tab:gui_throughput}
\footnotesize
\setlength{\tabcolsep}{4pt}
\resizebox{\textwidth}{!}{%
\begin{tabular}{@{}lccccc@{}}
\toprule
Setup & \shortstack{Rendering\\/capture} & \shortstack{Network\\/backend} & \shortstack{DOM\\/state parsing} & \shortstack{Single-env\\throughput (steps/s)} & \shortstack{Parallel-env\\throughput (steps/s)} \\
\midrule
Syn-Web (MiniWoB++, BrowserGym) & 0.5-2.0s & negligible–50 ms & 10–80 ms & 5–20 & 100–800 (64 envs) \\
Self-Web (WebArena, VWA) & 0.3–1.0 s & 0.1–0.8 s & 0.1–0.5 s & 0.5–2 & 15–150 (32--128 envs) \\
Desktop VM (OSWorld, ScreenAgent) & 0.5–2.0 s & var. file/app I/O & 0.1–1.0 s & 0.2–1.5 & 5–60 (16--64 VMs) \\
Mobile Emu (AndroidEnv, AndroidWorld) & 0.5–2.0 s & 0–1.0+ s & 0.1–0.5 s & 0.3–2 & 20–300 (32--256 emus) \\
WM/Latent rollout & bypassed & bypassed & light symbolic state & 10–100+ (sim) & $10^3$–$10^4$ (sim) \\
\bottomrule
\end{tabular}
}
\vspace{0.5em}
\parbox{\linewidth}{\scriptsize
Abbreviations: Syn-Web = synthetic browser tasks; Self-Web = self-hosted web benchmarks; VWA = VisualWebArena; Emu = emulator; WM = world model. Ranges denote order-of-magnitude planning values and exclude policy/model inference time. Columns are not strictly additive: components (rendering, network, parsing) run partially in pipeline, and throughput reflects end-to-end interaction rates under practical parallelism and caching. Representative sources: MiniWoB++ and CC-Net \citep{liu2018miniwob,humphreys2022ccnet}; WebArena, VisualWebArena, and BrowserGym \citep{zhou2023webarena,koh2024visualwebarena,chezelles2024browsergym}; OSWorld and ScreenAgent \citep{xie2024osworld,niu2024screenagent}; AndroidEnv and AndroidWorld \citep{toyama2021androidenv,rawles2024androidworld}; DreamGym and UI-Simulator \citep{chen2025dreamgym,wang2025uisim}.
}
\end{table}

For this reason, world models and latent-space training are likely to become more central rather than less. A promising long-term direction is to train agents that can alternate between two regimes: grounded interaction with the real interface, and accelerated imagination over learned interface dynamics. Such models would not replace real environments, because final success still depends on precise grounding in pixels, latency, and platform-specific behaviors. However, they could substantially reduce the cost of exploration, improve long-horizon credit assignment, and enable counterfactual reasoning about alternative action sequences before expensive execution. For GUI agents, this is a path to practical RL at scale; for computer-use agents, it is a step toward building internal models of how software ecosystems behave.

\subsubsection{Hierarchical Control}

Another recurring insight from this survey is that computer use is not a monolithic reasoning problem. The evidence reviewed in Section~\ref{sec:perception} shows that explicit deliberation can improve planning but degrade visual grounding, while Section~\ref{sec:memory} shows that long-horizon success depends on learned memory compression and retrieval rather than ever-longer context windows. Taken together, these findings point toward a hierarchical view of future agents: fast perceptual-action loops for local execution, slower reasoning modules for strategy shifts, and memory systems that preserve task-relevant state across long trajectories.

Here RL can play a broader role than optimizing action tokens alone. It can be used to learn \emph{when to think}, \emph{when to look closer}, \emph{when to retrieve memory}, and \emph{when to ask for help}. This kind of adaptive control is likely to matter even more for general computer-use agents than for current GUI benchmarks, because open environments contain a wider mixture of routine operations and rare high-stakes decisions. A mature computer-use agent should therefore optimize not only task reward, but also its allocation of attention, latency, and reasoning budget.

\subsection{Deployment and Governance}

Beyond capability, the long-term trajectory of computer-use agents depends on whether they can be deployed safely, evaluated realistically, and integrated into a governed digital ecosystem.

\subsubsection{Safety, Adaptation, and Evaluation}

The path toward digital inhabitants also raises a harder safety question. Current concerns already include phishing prompts, deceptive layouts, unsafe clicks, and irreversible operations~\citep{zhang2025attacking,kuntz2025harm}. But once agents become persistent and self-improving, the relevant risk is no longer a single bad action; it is the accumulation of miscalibrated behavior over time. Continual RL, replay-based adaptation, and cross-platform transfer are therefore double-edged: they are necessary for maintaining stable performance and environmental robustness under gradual interface drift, distribution shifts, and dynamic task changes, but they can also propagate unsafe behavioral shortcuts, amplify hidden vulnerability risks, or even gradually erase previously aligned safety guardrails and ethical constraints if not equipped with explicit regularization and strict boundary restrictions. Without well-designed constraint mechanisms and continual safety supervision, these adaptive learning paradigms will easily compromise model alignment and lead to unpredictable risky behaviors in open-ended real-world scenarios.

\rev{These limitations can be organized into three practical failure-mode groups:}
\begin{itemize}
    \item \rev{\textbf{Optimization failures.} Long-horizon error accumulation and reward hacking should be evaluated through horizon-stratified success, recovery rate, and adversarial reward tests.}
    \item \rev{\textbf{Evaluation failures.} LLM-as-judge unreliability, closed-source unverifiability, and reproducibility barriers call for judge-agreement checks, human-audited subsets, and explicit reporting of model/API versions, prompts, logs, random seeds, rollout budgets, and environment snapshots.}
    \item \rev{\textbf{Deployment failures.} Sim-to-real gaps and unsafe irreversible actions require transfer-degradation measurements, sandboxing, permission gates, rollback mechanisms, and human override protocols.}
\end{itemize}

This is why future evaluation must move beyond static benchmark success rates. For GUI agents, we need protocols that test reliability under interface updates, adversarial perturbations, partial failures, and recovery scenarios. For computer-use agents more broadly, the field will need benchmarks closer to digital operations than to isolated tasks: persistent identities, multi-application workflows, interrupt handling, permission boundaries, and human oversight at varying levels of granularity. In parallel, constrained RL and human-in-the-loop mechanisms should be treated as first-class components of the training objective rather than deployment-time patches. More broadly, safe deployment will require infrastructure-level guarantees in addition to policy-level alignment: identity systems that make agents legible, rule systems that specify their authority boundaries, and execution environments that support auditing, rollback, and accountability by default.

\subsubsection{Future Computer Use}

Taken together, these trends suggest that GUI agents are not merely one application area of RL, but a concrete route toward more general computer-use intelligence. They expose the full stack of difficulties that such systems must eventually solve: grounding in messy interfaces, acting under delayed and imperfect feedback, remembering long interaction histories, adapting to non-stationary software, and operating safely under real-world constraints. RL is unlikely to solve all of these challenges alone, but it is the framework that most naturally connects them through sequential optimization.

From the perspective of this survey, the distinctive opportunity is therefore not simply to make agents click more accurately or finish benchmarks more efficiently. It is to develop agents that can \emph{live in} digital environments in the same sense that modern language models can already \emph{speak in} natural language: persistently, adaptively, and under meaningful feedback. Yet the full realization of that vision may require a deeper transition, from agents operating human-oriented computers to agents inhabiting machine-oriented digital worlds. If that transition occurs, the study of reinforcement learning for GUI agents may ultimately be remembered not as a niche subfield, but as an early foundation for the broader science of digital inhabitants and agent-native infrastructure.

\section{Conclusion}

This survey provides a comprehensive analysis of Reinforcement Learning for GUI agents, covering offline, online, and hybrid paradigms alongside key dimensions like reward engineering, data efficiency, and technical innovations. Three core findings emerge from this landscape. First, the lack of easily readable rewards in GUI environments forces agents to interpret multimodal evidence, driving the adoption of hybrid reward schemes. Second, severe I/O latency makes data efficiency a binding constraint, motivating the use of latent-space world models over on-policy algorithms. \rev{Third, early evidence suggests that structured action spaces and rewards may reduce reliance on explicit reasoning annotations, while balancing fast intuitive grounding with slow deliberative planning remains a fundamental challenge.}

Looking ahead, future research will likely focus on process reward models for reasoning trajectories, continual learning for interface updates, cross-platform agents, and dynamic cognitive architectures. As agents inevitably transition to production, establishing formal safety guarantees and real-world deployment benchmarks becomes indispensable. More importantly, the long-term trajectory of the field may extend beyond training agents to operate human-oriented interfaces more effectively. If computer-use agents are to become genuine digital inhabitants, they will likely require not only stronger policies, but also agent-native infrastructure: persistent identities, explicit authority boundaries, auditable execution environments, and operating substrates designed for machine actors. Ultimately, the convergence of reasoning-enhanced VLMs with hybrid RL training signals a fundamental shift in intelligent digital interaction, where RL will remain central to addressing sequential decision-making under uncertainty. We hope this survey inspires further advances not only toward robust, generalizable, and safe GUI agents, but also toward consolidating and expanding the broader theoretical and technical foundations of digital inhabitants.

\bibliography{main}
\bibliographystyle{tmlr}

\addtocontents{toc}{\protect\setcounter{tocdepth}{-1}}
\appendix
\begingroup
\section{Survey Methodology}
\label{app:survey-methodology}

This appendix documents how we construct the literature corpus used in this
survey.

\subsection{Search Protocol}

We constructed the corpus through keyword search, venue tracking, prior-survey
comparison, and backward/forward citation inspection. The search sources include
Google Scholar, arXiv, Semantic Scholar, ACL Anthology, OpenReview, and the
proceedings or accepted-paper lists of major AI, NLP, vision, RL, systems, and
HCI venues. We also tracked benchmark leaderboards, model reports, and technical
reports when they introduced systems or environments that are directly analyzed
in the survey.

The query set combined GUI-agent terms, platform terms, and RL-specific terms,
including: ``GUI agents'', ``computer-use agents'', ``computer control agents'',
``web agents'', ``mobile GUI agents'', ``desktop agents'', ``OS agents'',
``visual grounding GUI'', ``UI grounding'', ``GUI automation benchmark'',
``reinforcement learning GUI agents'', ``RL for web agents'', ``online RL GUI
agents'', ``offline RL GUI agents'', ``preference optimization GUI agents'',
``GRPO GUI agents'', ``RLVR GUI agents'', ``reward model GUI agents'', ``world
model web agents'', and ``continual learning GUI agents''. The working cutoff
for the current manuscript snapshot is August, 2026.

\subsection{Inclusion and Exclusion Criteria}

We include works that satisfy at least one of the following criteria: (i) they
study agents that perceive and act through graphical, web, mobile, desktop, or
operating-system interfaces; (ii) they introduce benchmarks, environments,
datasets, or interaction protocols for GUI/web/mobile/desktop agents; (iii) they
apply reinforcement learning, preference optimization, verifiable rewards,
online self-improvement, model-based rollouts, or related policy-optimization
methods to GUI-agent training; or (iv) they introduce reward, grounding,
memory, planning, exploration, or infrastructure mechanisms that materially
affect RL-style GUI-agent training.

We exclude works that only mention GUIs as incidental front ends, papers on
general RL or LLM alignment without a GUI-agent connection, purely static UI
understanding papers without agentic decision making unless they provide
datasets or grounding signals used by GUI agents, generic RPA/Selenium-style
automation without learning-based agents, and blog posts or product pages unless
they describe influential frontier systems that shape the field's empirical
baseline. General RL, LLM, VLM, and safety references cited for background are
not counted as part of the core GUI-agent taxonomy corpus.

\subsection{Preprint and Version Handling}

Because GUI-agent research changes quickly, we include arXiv and OpenReview
preprints when they introduce relevant methods, benchmarks, datasets, or
empirical results. When a paper has both a preprint and a peer-reviewed version,
we cite the peer-reviewed version when available and use the preprint for
details that are absent from the proceedings version.

\subsection{Taxonomy Assignment}

Each included work is assigned according to the dominant role it plays in the
survey. For method papers, the primary coding dimensions are learning regime
(offline, online, hybrid, preference-based, model-based, or curriculum-based),
interaction regime (static dataset, simulated environment, live environment, or
mixed), action space (coordinate-level, UI-element-level, code/API-level, or
hybrid), reward source (rule/verifier, environment state, human preference,
LLM-as-judge, learned reward model, or composite reward), data source
(demonstrations, synthetic trajectories, self-generated rollouts, human
feedback, AI feedback, or mixed data), environment type (web, mobile, desktop,
OS-level, or cross-platform), and evaluation setting (benchmark, metric, model
size, modalities, rollout budget, and implementation status when available).
Benchmark and dataset papers are coded by platform, observation/action
interface, reward or verification availability, task horizon, reproducibility,
and whether they support offline training, online evaluation, or both.

If a work spans multiple categories, we assign it to the category that explains
its main contribution in this survey and record secondary tags separately. For
example, a system that combines supervised initialization with online
fine-tuning is placed under hybrid RL if the offline-to-online transition is the
central contribution, while a grounding-focused work with an RL reward is placed
under specialized capabilities if the contribution is primarily visual
localization rather than a general training pipeline.

\begin{table}[t]
\centering
\small
\caption{Core taxonomy snapshot used for this survey. Commercial technical
releases are tracked as frontier-system records, not as peer-reviewed method
papers unless accompanied by archival publications.}
\label{tab:core-taxonomy-snapshot}
\begin{tabular}{p{0.24\linewidth}p{0.68\linewidth}}
\toprule
\textbf{Taxonomy group} & \textbf{Representative records in the current
snapshot} \\
\midrule
Foundational web-GUI systems &
MiniWoB~\citep{shi2017world}, MiniWoB++~\citep{liu2018miniwob},
WebN-T5~\citep{gur2023understanding}, WebShop~\citep{yao2022webshop},
WebGPT~\citep{nakano2021webgpt}. \\
Benchmarks and environments &
WebArena~\citep{zhou2023webarena}, Windows Agent
Arena~\citep{bonatti2024windows}, OSWorld~\citep{xie2024osworld},
Android-in-the-Wild~\citep{rawles2023androidinthewild},
ScreenSpot-Pro~\citep{li2025screenspot}, OS-Atlas~\citep{wu2024atlas},
WebVoyager~\citep{he2024webvoyager}, UI-Vision~\citep{nayak2025ui},
AndroidWorld~\citep{rawles2024androidworld},
GUI-Odyssey~\citep{lu2025guiodyssey}. \\
Frontier GUI-agent systems &
Claude Computer Use~\citep{anthropicIntroducingComputer}, OpenAI
CUA~\citep{openaiComputerUsingAgent}, Gemini 2.5
CUA~\citep{geminiComputerUsingAgent}, UI-TARS~\citep{qin2025ui,wang2025ui},
MAI-UI~\citep{zhou2025mai}, DART-GUI~\citep{li2025efficient},
Mano~\citep{fu2025mano}, GELab-Zero~\citep{gelab_engine},
Falcon-UI~\citep{shen2024falcon}. \\
RL optimization methods &
Digi-Q~\citep{bai2025digi}, UI-TARS ARPO~\citep{lu2025arpo},
Agent Q~\citep{putta2024agent}, WebRL~\citep{qi2024webrl},
MobileRL~\citep{xu2025mobilerl}, Explorer~\citep{pahuja2025explorer},
WebAgent-R1~\citep{wei2025webagent}, DigiRL~\citep{bai2024digirl},
UI-S1~\citep{lu2025ui}, Hi-Agent~\citep{wu2025hi},
DynaWeb~\citep{ding2026dynaweb}, UltraCUA~\citep{yang2025ultracua}. \\
Specialized capabilities &
InfiGUI-G1~\citep{liu2026infiguig1}, GUI-Eyes~\citep{chen2026guieyes},
SE-GUI~\citep{yuan2025enhancing}, GUI-G\textsuperscript{2}~\citep{tang2025guig2},
UI-AGILE~\citep{lian2025ui}, UI-R1~\citep{lu2025uir1},
GUI-R1~\citep{luo2025guir1}, VSC-RL~\citep{wu2025vsc},
BacktrackAgent~\citep{wu2025backtrackagent}. \\
\bottomrule
\end{tabular}
\end{table}

\endgroup

\end{document}